\documentclass[10pt,twocolumn,letterpaper]{article}

\usepackage[pagenumbers]{cvpr} 
\usepackage{booktabs}    
\usepackage{multirow}    
\usepackage{makecell}    
\usepackage{graphicx}    
\usepackage{algorithm}
\usepackage{algpseudocode}
\usepackage[misc]{ifsym}
\usepackage[hang]{footmisc}

\definecolor{cvprblue}{rgb}{0.21,0.49,0.74}
\usepackage[pagebackref,breaklinks,colorlinks,allcolors=cvprblue]{hyperref}

\makeatletter
\newcommand{\removelatexerror}{\let\@latex@error\@gobble}
\makeatother

\newcommand\blfootnote[1]{
    \begingroup
    \renewcommand\thefootnote{}\footnote{#1}
    \addtocounter{footnote}{-1}
    \endgroup
}

\def\paperID{2150} 
\def\confName{CVPR}
\def\confYear{2026}

\title{SMRABooth: Subject and Motion Representation Alignment for Customized Video Generation}

\author{Xuancheng Xu\textsuperscript{1} \quad 
Yaning Li\textsuperscript{1} \quad
Sisi You\textsuperscript{1,$\dagger$} \quad
Bing-Kun Bao\textsuperscript{1,2} \\
\textsuperscript{1}Nanjing University of Posts and Telecommunications \quad  
\textsuperscript{2}Peng Cheng Laboratory \\
{\tt\small 2024010131@njupt.edu.cn} \quad
{\tt\small liyaning@njupt.edu.cn} \\
{\tt\small ssyou@njupt.edu.cn} \quad
{\tt\small bingkunbao@njupt.edu.cn} \\
\\
\vspace{1em}
Project page: \url{https://smrabooth.github.io}
}

\begin{document}

\twocolumn[{
    \maketitle
    \begin{center}
        \vspace{-2.5em}
        \centering
        \includegraphics[width=1.0\textwidth]{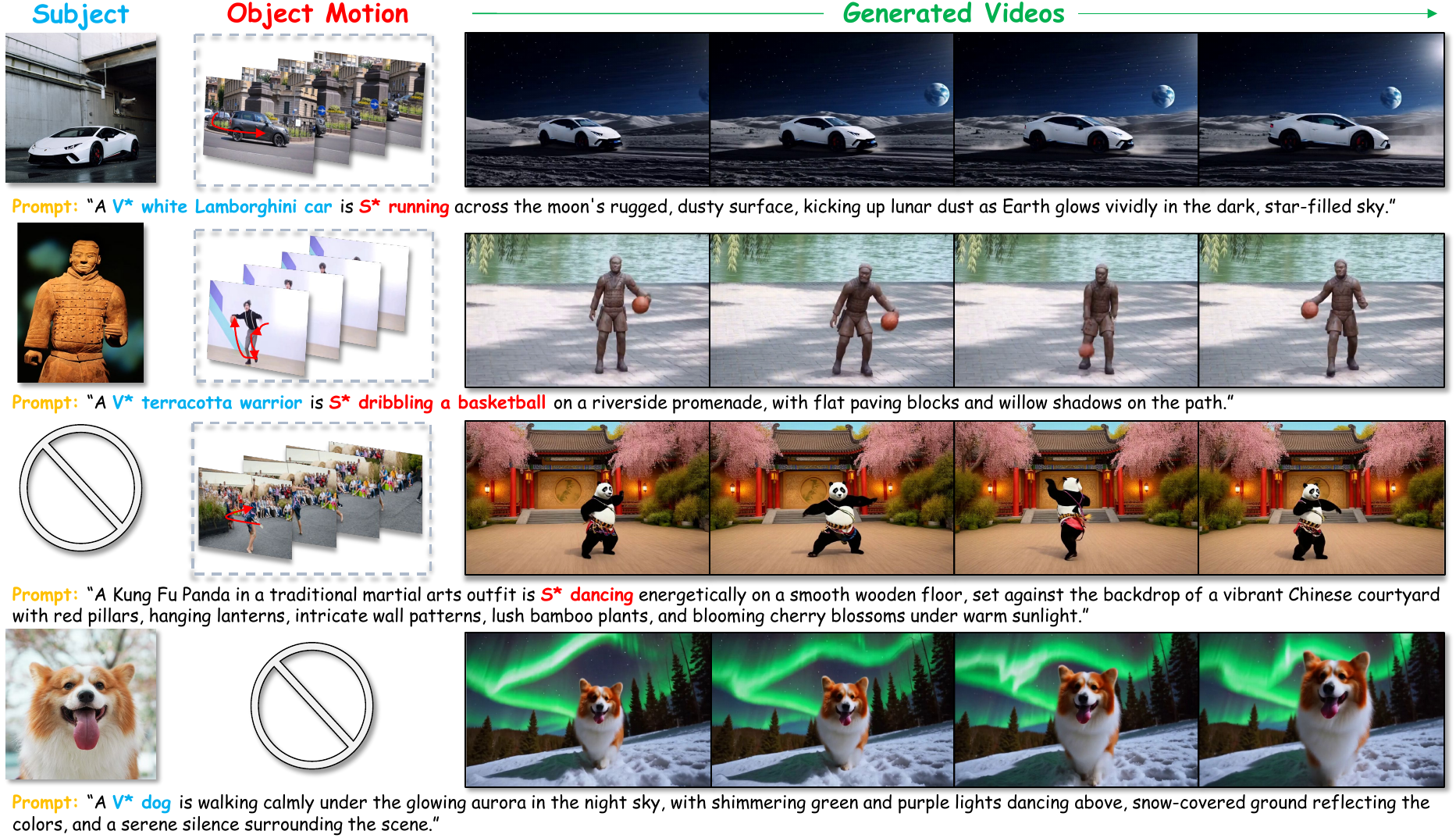}
        \vspace{-2em}
        \captionof{figure}{SMRABooth is a text-to-video customized generation framework that allows users to individually or jointly control the subject and object motion. Our SMRABooth can handle with multi types of motion, including regular linear motion, irregular curvilinear motion, rotational motion, and conceptual motion involving certain sports and musical instruments.}
        \label{teaser}
    \end{center}
    \vspace{-1em}
}]

\blfootnote{$^\dagger$ Corresponding Author}

\begin{abstract}
Customized video generation aims to produce videos that faithfully preserve the subject's appearance from reference images while maintaining temporally consistent motion from reference videos. 
Existing methods struggle to ensure both subject appearance similarity and motion pattern consistency due to the lack of object-level guidance for subject and motion.
To address this, we propose SMRABooth, which leverages the self-supervised encoder and optical flow encoder to provide object-level subject and motion representations. These representations are aligned with the model during the LoRA fine-tuning process.
Our approach is structured in three core stages:
(1) We exploit subject representations via a self-supervised encoder to guide subject alignment, enabling the model to capture overall structure of subject and enhance high-level semantic consistency.
(2) We utilize motion representations from an optical flow encoder to capture structurally coherent and object-level motion trajectories independent of appearance.
(3) We propose a subject-motion association decoupling strategy that applies sparse LoRAs injection across both locations and timing, effectively reducing interference between subject and motion LoRAs.
Extensive experiments show that SMRABooth excels in subject and motion customization, maintaining consistent subject appearance and motion patterns, proving its effectiveness in controllable text-to-video generation. 
\end{abstract}    
\section{Introduction}
Customized video generation based on Text-to-Video (T2V) models~\cite{blattmann2023stable, hongcogvideo, chen2024videocrafter2} aims to preserve the subject's appearance from reference images while ensuring temporally consistent motion from reference videos (e.g., Fig.~\ref{teaser}).  
Most models ~\cite{wei2024dreamvideo, zhao2024motiondirector, bi2025customttt} require two-stage training to separately extract subject and motion features, which are then combined during inference to achieve customized generation.

\begin{figure}[t]
    \centering
    \includegraphics[width=0.48\textwidth]{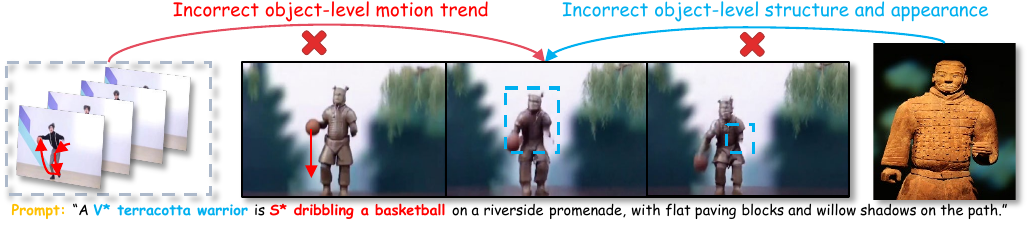}
    \vspace{-2em}
    \caption{Existing methods lack object-level information, leading to poor motion trends and incomplete subject preservation due to limited structural awareness.}
    \label{fig_badcase}
    \vspace{-0.7cm}
\end{figure}

Existing methods~\cite{wei2024dreamvideo, wumotionbooth, jiang2024videobooth, zhao2024motiondirector, bi2025customttt, wang2025dualreal} have made significant progress in generating videos with consistent subject identities and motion patterns.
However, the failure to incorporate object-level information, including the holistic structure of the subject and the overall motion trends, results in the generalized subject appearance and the overall motion patterns still being unsatisfactory (e.g., the incorrect human-basketball motion trends and the missing hands of people in Fig.~\ref{fig_badcase}).

Besides, previous U-Net-based methods ~\cite{wei2024dreamvideo, zhao2024motiondirector, bi2025customttt} separately fine-tune subject and motion LoRAs on spatial and temporal layers to reduce the impact of combining during inference.
However, DiT-based backbones~\cite{yang2024cogvideox, wan2025wan} lack a clear distinction between spatial and temporal layers, leading to interference when they are directly combined during the inference phase. This causes the generated video to overly rely on the entangled features, resulting in artifacts and poor-quality outputs.

To address these issues, we propose SMRABooth, a stage-by-stage framework for customized video generation.
First, for subject learning, we introduce the \textbf{Su}bject \textbf{R}epresentation \textbf{A}lignment (\textbf{SuRA}) module.  
Motivated by the strong correlation between a model's features and external vision encoder representations~\cite{yu2024representation}, SuRA leverages a self-supervised vision encoder to utilize its pre-trained ability to model subject representations, enhancing global semantic understanding and spatial structure awareness through feature alignment.
By training a subject low-rank matrix (LoRA) under the supervision of an external vision encoder, we accurately preserve the subject’s appearance from reference images.

Secondly, to address the limitation in capturing object-level motion, we introduce the \textbf{Mo}tion \textbf{R}epresentation \textbf{A}lignment (\textbf{MoRA}) module for motion learning.
MoRA addresses this by leveraging an optical flow encoder to explicitly extract temporal motion representation from reference videos in the pixel space. 
This encoder excels at modeling absolute trajectory changes while filtering out appearance-related redundancies, providing a clear signal of object motion patterns.
By aligning the extracted temporal motion representations with the diffusion model, MoRA captures coherent motion trajectories and trains a motion LoRA to enforce fine-grained temporal consistency.

To address the entanglement of spatial and temporal layers in DiT-based backbone, which couples appearance and motion patterns, we propose a subject-motion association decoupling strategy.
The strategy is motivated by the observed sparsity of LoRAs in both injection locations and timing through our LoRA sparsity experiments. 
Specifically, for injection locations, although the spatial and temporal layers in DiT are coupled, we find that different linear layers have varying degrees of influence on the subject and motion.
Our findings indicate that fine-tuning a few layers for subject and motion LoRAs can achieve almost similar performance as full-layer LoRA fine-tuning while effectively addressing the coupling issue between LoRAs.
Moreover, for injection timing, the early stages focus on shaping motion, whereas the later stages refine appearance.
Thus we apply lower subject LoRA weights before a critical point in the generation process, followed by higher weights afterward, effectively aligns with the coarse-to-fine nature of the timing.  
By balancing appearance fidelity and video coherence, this strategy ensures high-quality video generation without the entanglement of subject and motion.

\begin{figure*}[!h]
    \centering
    \includegraphics[width=0.95\textwidth]{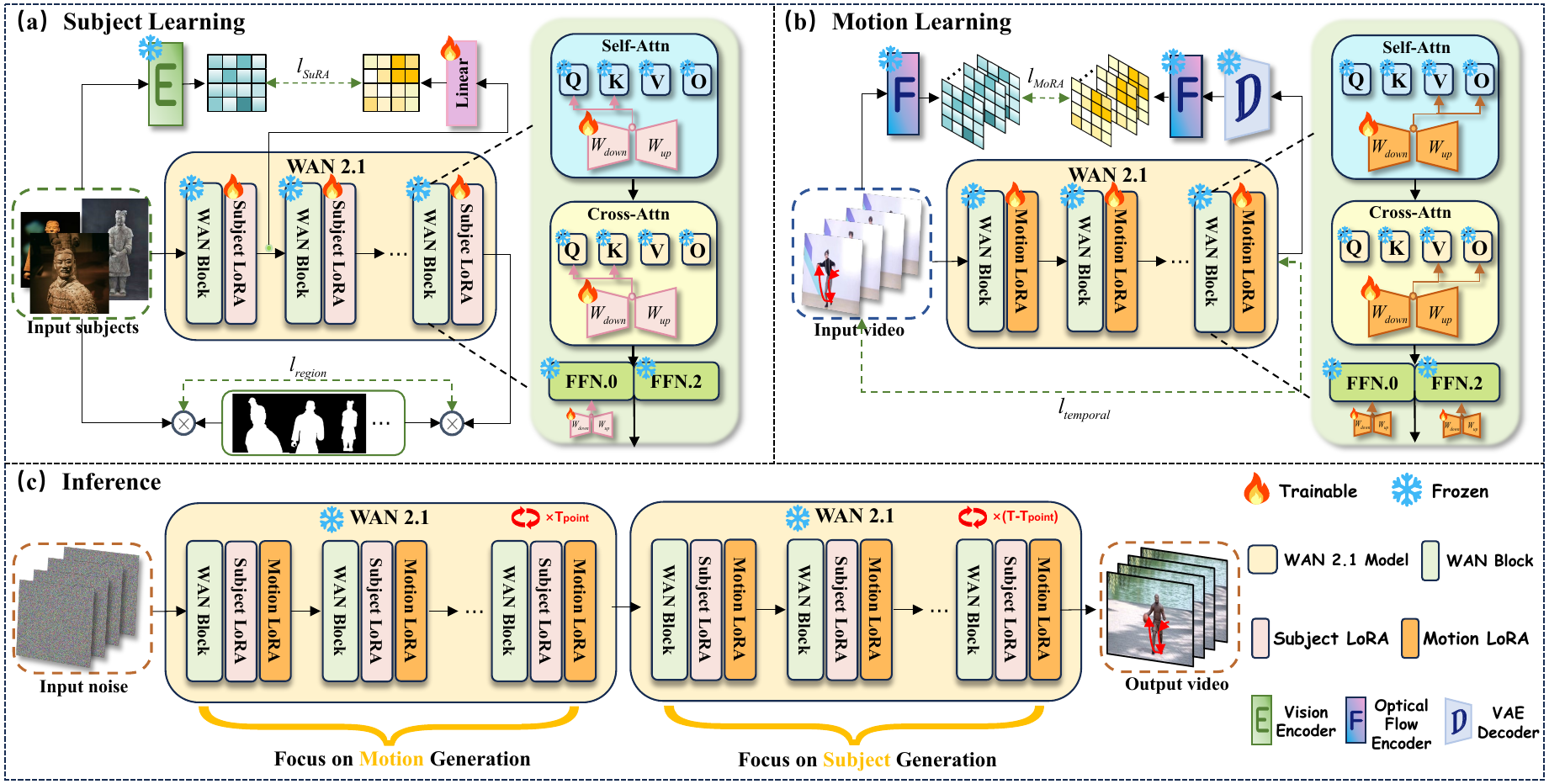}
    \vspace{-0.35cm}
    \caption{\textbf{Overview of SMRABooth.} The framework splits customized video generation into two stages: subject learning and motion learning. Subject learning aligns global spatial features from the vision encoder to enhance fidelity, while motion learning utilizes temporal motion representations from the optical flow encoder to guide motion generation. The pretrained video diffusion model remains frozen during training, and LoRAs are merged at inference to generate customized videos. For simplicity, text input is omitted from the figure. V* and S* are specific tokens used to represent subject and motion without intrinsic meanings.}
    \label{fig_pipeline_dit}
    \vspace{-0.5cm}
\end{figure*}
Overall, our contribution can be summarized as:
\begin{itemize}
\item \textcolor{black}{We design \textbf{separate subject and motion representation alignment} modules to guide the model in aligning with external object-level representations during LoRA fine-tuning, enhancing subject fidelity and motion coherence.} 

\item \textcolor{black}{We propose a novel \textbf{subject-motion association decoupling strategy} that leverages the sparse characteristics of LoRAs in injection locations and timing to balance subject and motion consistency in video generation.}

\item \textcolor{black}{Our unified representation alignment framework enables stage-by-stage customization of subject and motion \textbf{across both DiT and U-Net based backbones}, achieving state-of-the-art fidelity in appearance preservation and motion consistency.}

\end{itemize}

\section{Related Works}  
\subsection{Text-to-Video Generation}  
Diffusion-based models have achieved remarkable success in generating high-quality images~\cite{songdenoising, ho2020denoising, rombach2022high, tao2022df, tao2023galip, podellsdxl} and videos~\cite{blattmann2023stable, hongcogvideo, singermake, khachatryan2023text2video, wu2023tune, yang2024cogvideox,liu2025javisdit} from text prompts.  
Methods like ~\cite{blattmann2023stable, zhou2022magicvideo} leverage spatial-temporal U-Nets, while ~\cite{guoanimatediff} introduces motion LoRA to customize temporal layers in T2I models.  
Foundation models~\cite{chen2023videocrafter1, chen2024videocrafter2, wang2023modelscope, wang2024lavie, Zeroscope}, trained on large-scale datasets~\cite{schuhmann2022laion, Bain21, chen2024panda}, exhibit strong generative capabilities. 
Recently, models like ~\cite{yang2024cogvideox, wan2025wan} have demonstrated the DiT~\cite{peebles2023scalable} backbone's strong capabilities in enhancing appearance and motion quality for video generation. Open-source models have further fueled enthusiasm, empowering users to create both realistic and imaginative videos while enabling private model customization.
\subsection{Customized Generation}  
Customized generation ~\cite{li2025comprehensive} integrates subject-specific features and controllable motion into image and video generation, ensuring fidelity to reference inputs and precise representation.  
For image generation, models like ~\cite{ruiz2023dreambooth, galimage, zhang2023prospect, 11209272, wu2024infinite} use special tokens to accurately represent subject information.  
In video generation, methods such as ~\cite{zhao2024magdiff,jiang2024videobooth,wumotionbooth} use mask to isolate subjects and minimize background interference, while~\cite{wei2024dreamvideo,wu2025customcrafter, huang2025dive, huang2025videomage} leverage spatial low-rank structures to enhance geometry preservation and detail rendering, ensuring fidelity and visual coherence.

For controllable motion generation, methods aim to guide video movements with temporal consistency and accurate motion representation.  
~\cite{wumotionbooth,yang2024direct} constrain attention to bounding box regions during inference, achieving coarse alignment but lacking fine-grained control.  
~\cite{wei2024dreamvideo} enhances motion alignment through temporal attention and reweighted Diffusion Loss.  
~\cite{wang2024motionctrl} incorporates camera parameters into temporal attention and introduces motion information using a T2I-Adapter~\cite{mou2024t2i} structure.  
Low-rank matrix fine-tuning, employed in~\cite{guoanimatediff,zhao2024motiondirector,wei2024dreamvideo, zhang2025flexiact,wang2025motion,zhang2025motioncrafter,bi2025customttt}, utilizes temporal LoRA~\cite{hulora} or Adapters~\cite{rebuffi2017learning} to learn temporal features from video data. 
Methods such as ~\cite{wang2025dualreal,chen2025jointtuner,wei2025dreamrelation} combine subject-motion LoRA groups to effectively learn appearance and motion information from datasets. This approach is applied to DiT-based backbones, which lack an inherent distinction between spatial and temporal layers.
These techniques collectively enhance customized generation by combining subject fidelity with precise motion control and robust temporal consistency.
\section{Methods}\label{sec:method}  
In this section, we present SMRABooth, a stage-by-stage framework for customized video generation based on subject and motion representation alignment with video diffusion models. The overall pipeline of SMRABooth is shown in Figure~\ref{fig_pipeline_dit}.  
Specifically, Sec.~\ref{sub:Subject} introduces the subject representation alignment module, which leverages global structure and semantics from reference images to enhance subject fidelity and coherence.  
Sec.~\ref{sub:Motion} describes the motion representation alignment module, capturing object-level motion trajectories while disentangling appearance from motion to improve motion similarity with reference videos.  
Finally, Sec.~\ref{sub:Inject} introduces a subject-motion association decoupling strategy. By leveraging the sparse characteristics of LoRAs in injection locations and timing, this method balances appearance fidelity and video coherence.
\noindent\subsection{Preliminary}\label{sub:Preliminary}
\textbf{Video Diffusion Transformer Model.}  
Text-to-video diffusion transformers, leveraging flow matching denoising strategies~\cite{lipman2022flow}, enable high-quality video generation. WAN2.1~\cite{wan2025wan} integrates a WAN block with flow matching to iteratively denoise Gaussian noise based on textual prompts.  
A pre-trained 3D VAE~\cite{kingma2013auto} encodes video data $X$ into latent representations $z = \mathcal{E}(X)$ via encoder $\mathcal{E}$, while a decoder $\mathcal{D}$ reconstructs frames as $X \approx \mathcal{D}(z)$. The WAN block employs self-attention for temporal-spatial dependencies and cross-attention for textual conditioning.  
The model is trained with conditioning from a text encoder $c_{txt}$ and optimized using a velocity prediction loss:
\begin{equation}\label{eqn-1}
\mathcal{L} = \mathbb{E}_{z_0, z_1, c_{txt}, t} \left\| u(z_t, c_{txt}, t; \theta) - v_t \right\|^2,
\end{equation}
where the interpolation $z_t$ between $z_0$ and $z_1$ is given by:  
\begin{equation}\label{eqn-2}
z_t = (1-t)z_0 + tz_1,
\end{equation}
and the velocity field indicates the flow direction:  
\begin{equation}\label{eqn-3}
v_t = z_1 - z_0.
\end{equation}

\noindent\textbf{Optical Flow.}  
Optical flow describes pixel-wise motion between consecutive video frames as motion vectors encoding direction and velocity. Given two frames $x_t$ at time $t$ and $x_{t+dt}$ at $t+dt$, optical flow is computed by estimating pixel displacements $(u, v)$ while preserving spatial smoothness.  
Recent optical flow encoders~\cite{ling2023learning,wang2024sea,ling2024scaleflow++,ling2024adfactory} estimate the optical flow field:
\begin{equation}\label{eqn-4}
\mathbf{F_{t,t+dt}} = F(x_t, x_{t+dt}),
\end{equation}
where $\mathbf{F_{t,t+dt}} \in \mathbb{R}^{H \times W \times 2}$ represents pixel-wise motion vectors. These encoders ensure precise motion estimation by leveraging advanced neural architectures.
\subsection{Subject Representation Alignment}\label{sub:Subject}  
Recent study~\cite{yu2024representation} has shown that aligning model features with external vision encoder representations enhances global semantic understanding and spatial structure awareness.
Building on this, we introduce subject representation alignment module to achieve global spatial control in subject customization.  
We employ a self-supervised vision encoder to extract subject representations. The spatial representation preserves accurate global spatial relationships between object parts while retaining rich semantic information, serving as the alignment target.
By calculating a cosine similarity loss, the target representations guide the learning process of intermediate features in the DiT blocks, enhancing the representational capacity of the model.
For customized subject generation, similarity loss is further employed to train subject LoRAs. 

Specifically, at each epoch, we sequentially select an image $I_{sub}^i$ from the provided set of subject images $I_{sub}$ and train the model to reconstruct it.  
To obtain robust global target representations, we incorporate a frozen DINOv2-ViT encoder $E$~\cite{oquab2023dinov2,darcet2023vitneedreg}, which extracts target features $\mathbf{y^*}$.
Specifically, $\mathbf{y^*} = E(I_{sub}^i) \in \mathbb{R}^{N \times D}$ represents the patch embeddings of the selected image, where $N$ denotes the number of patches, and $D$ is the embedding dimension. 
These target features provide a high-dimensional representation of the input image, encapsulating rich semantic and spatial information for subsequent model training.  
During training, we extract the denoised features $z_{t}^{1}$ after the first transformer layer at timestep $t$.
Then we introduce a trainable multilayer perceptron (MLP), denoted as $h_{\phi}$, to align the feature dimensions of $z_{t}^{1}$ with those of the target features $\mathbf{y^*}$.  
The spatial representation alignment module allows the intermediate features to approximate the global spatial structure and high-level semantic features of the target features, significantly improving the quality and fidelity of the generated subject. Feature alignment is achieved through the following loss function, defined as:
\begin{equation}\label{eqn-5}
\mathcal{L}_{\mathrm{SuRA}}(\theta) = -\mathbb{E}_{\mathbf{z},\boldsymbol{v},t}\left[\frac{1}{N}\sum_{n=1}^{N}\frac{\mathbf{y}^{*[n]} \cdot h_{\phi}(z_t^{1[n]})}{\|\mathbf{y}^{*[n]}\| \cdot \|h_{\phi}(z_t^{1[n]})\|}\right],
\end{equation}
where $N$ denotes the total number of patches.
Moreover, to prevent spatial LoRA from overfitting to the background of the subject image during training, we introduce masks $\mathbf{M}$ generated by SAM~\cite{kirillov2023segment}, which enforce the model to focus only on the subject region. 
Formally, the masked velocity prediction loss is defined as:
\begin{equation}\label{eqn-6}
\mathcal{L}_{\mathrm{region}} = \mathbb{E}_{z_0, z_1, c_{txt}, t} \left\| u(z_t, c_{txt}, t; \theta)\cdot \mathbf{M} - v_t\cdot \mathbf{M} \right\|^2,
\end{equation}
where $M$ represents the mask applied to the subject region. 
During training, the overall loss function is defined as:
\begin{equation}\label{eqn-7}
\mathcal{L} = \mathcal{L}_\mathrm{region} + \lambda \mathcal{L}_{\mathrm{SuRA}},
\end{equation}
where $\lambda > 0$ is a hyperparameter that balances the subject representation alignment loss.
\subsection{Motion Representation Alignment}\label{sub:Motion}
It is well known that optical flow encoders effectively capture global, object-level temporal motion by modeling trajectory changes while filtering out irrelevant appearance information~\cite{chefer2025videojam}.
Leveraging this capability, we introduce the motion representation alignment module to enhance motion learning by aligning with global movement trends while filtering out irrelevant appearance information, enabling more effective motion customization. 
We adopt SEA-RAFT~\cite{wang2024sea}, denoted as $\mathbf{F}$, as our optical flow encoder due to its robust performance in extracting pixel motion from real reference videos $X = \{x_i\}_{i=1}^N$. As shown in Eq.~\ref{eqn-4}, the real motion features are calculated as:
\begin{equation}\label{eqn-8}
\begin{aligned}
    \mathbf{F_{\{1,N\}}} &= \{F(x_1, x_2), F(x_2, x_3), \dots, F(x_{N-1}, x_N)\},
\end{aligned}
\end{equation}
where $\mathbf{F_{\{1,N\}}} \in \mathbb{R}^{{(N-1)} \times H \times W \times 2}$ represents pixel-wise horizontal and vertical motion across all adjacent frames of the reference video.
After completing denoising at each time step, we restore the video features $Z = \left\{z_j\right\}_{j=1}^{(N-1)/{4} +1}$ from latent space to pixel space using the inverse transform of Eq.~\ref{eqn-2} at timestep $t$:
\begin{equation}\label{eqn-9}
\{\widetilde{x}_i\}_{i=1}^N = \{\mathbf{\mathcal{D}}(z_{t,j}-t * u(z_{t,j}, c_{txt}, t; \theta))\}_{j=1}^{(N-1)/{4} +1},
\end{equation}
where $\mathbf{\mathcal{D}}$ represents the pre-trained 3D VAE decoder. 
Using the calculation method as in Eq.~\ref{eqn-7}, we extract the motion features from the denoised video:
\begin{equation}\label{eqn-10}
\begin{aligned}
    \mathbf{\widetilde{F}_{\{1,N\}}} &= \{F(\widetilde{x}_1, \widetilde{x}_2), F(\widetilde{x}_2, \widetilde{x}_3), \dots, F(\widetilde{x}_{N-1}, \widetilde{x}_N)\},
\end{aligned}
\end{equation}
where $\mathbf{\widetilde{F}_{\{1,N\}}} \in \mathbb{R}^{{(N-1)} \times H \times W \times 2}$ represents pixel-wise horizontal and vertical motion across all adjacent frames of the denoised video.
Finally, motion feature alignment is realized through the L1 loss, defined as:
\begin{equation}\label{eqn-11}
\mathcal{L}_{MoRA}=||\mathbf{F_{\{1,N\}}}-\mathbf{\widetilde{F}_{\{1,N\}}}||,
\end{equation}

During training, we apply temporal LoRAs into the WAN Block to update frame correlations using the temporal velocity prediction loss $\mathcal{L}_\mathrm{temporal}$:
\begin{equation}\label{eqn-12}
\begin{aligned}
    \mathcal{L}_\mathrm{temporal} = \mathbb{E}_{z_0, z_1, c_{txt}, t} \left\| u(z_t, c_{txt}, t; \theta) - v_t \right\|^2,
\end{aligned}
\end{equation}
and the overall loss function is defined as:
\begin{equation}\label{eqn-13}
\mathcal{L} = \mathcal{L}_\mathrm{temporal} + \alpha \mathcal{L}_{\mathrm{MoRA}},
\end{equation}
where $\alpha>0$ is a hyperparameter that balances the contributions of different loss terms.
\subsection{Subject-Motion Association Decoupling}\label{sub:Inject}
Unlike U-Net-based backbones~\cite{Zeroscope,wang2023modelscope}, DiT-based backbones like WAN2.1~\cite{wan2025wan} lack spatial-temporal layer disentanglement. Injecting both subject and motion LoRAs into all layers disrupts the subject-motion balance, leading to artifacts, poor video quality, and excessive background copying, failing to follow prompts (e.g., Table~\ref{table_ablation_dit}, Combination\textcircled{1}).
\begin{figure}[t]
    \centering
    \includegraphics[width=0.48\textwidth]{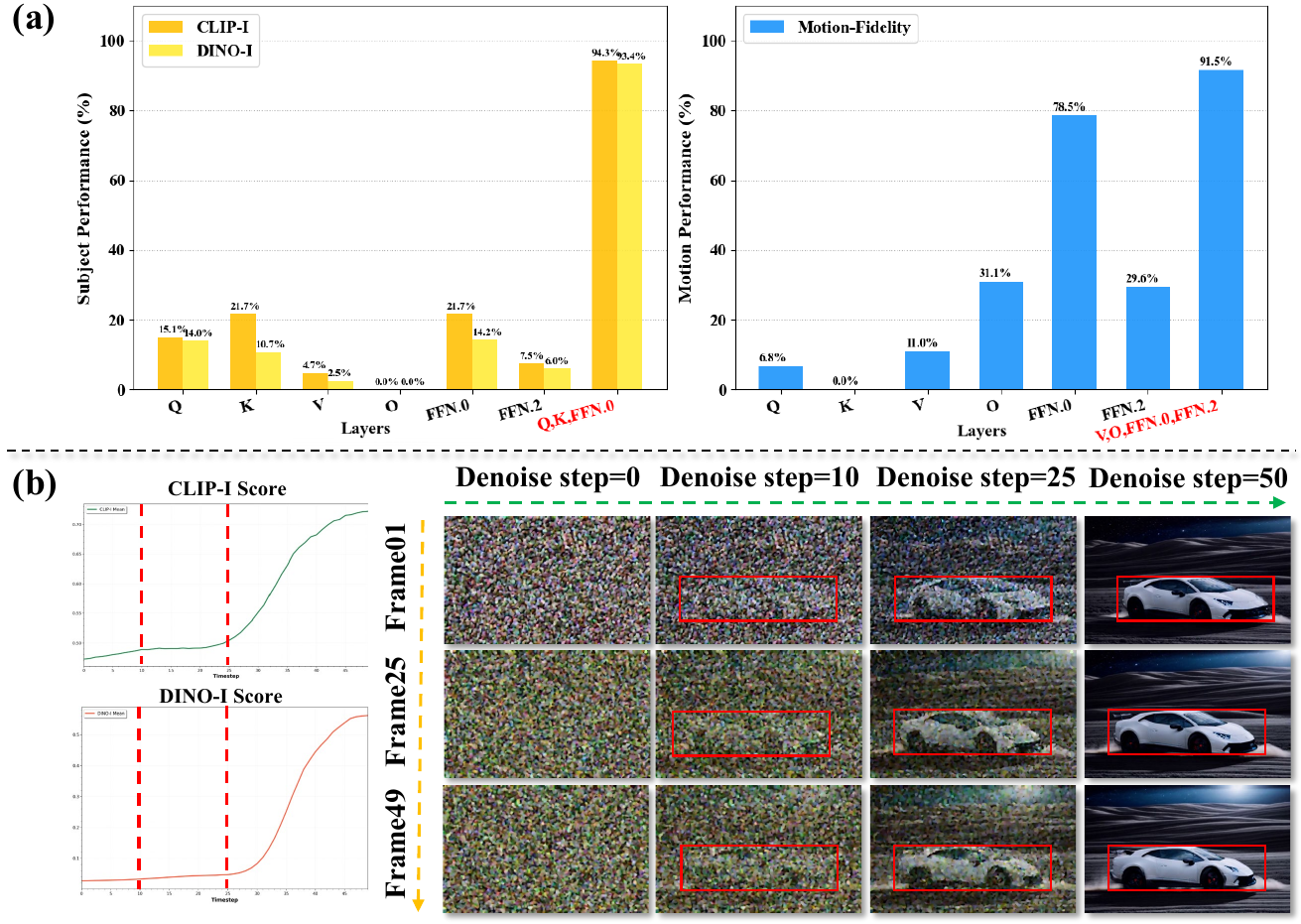}
    \vspace{-2em}
    \caption{(a) The bar charts present the sparse experiment results, highlighting the effectiveness of different layers for subject and motion learning. In the yellow bar chart, we apply the subject LoRA during inference to test its effectiveness in preserving subject similarity. In the blue bar chart, we apply the motion LoRA to evaluate its ability to maintain motion coherence. (b) The left table shows changes in CLIP-I and DINO-I metrics, indicating that subject details are refined between 10 and 25 denoising steps. The right figure illustrates the denoising process, where motion develops early, and the subject’s appearance emerges later.}
    \label{fig_timestep}
    \vspace{-0.7cm}
\end{figure}
\begin{figure*}[t]
    \centering
    \includegraphics[width=0.95\textwidth]{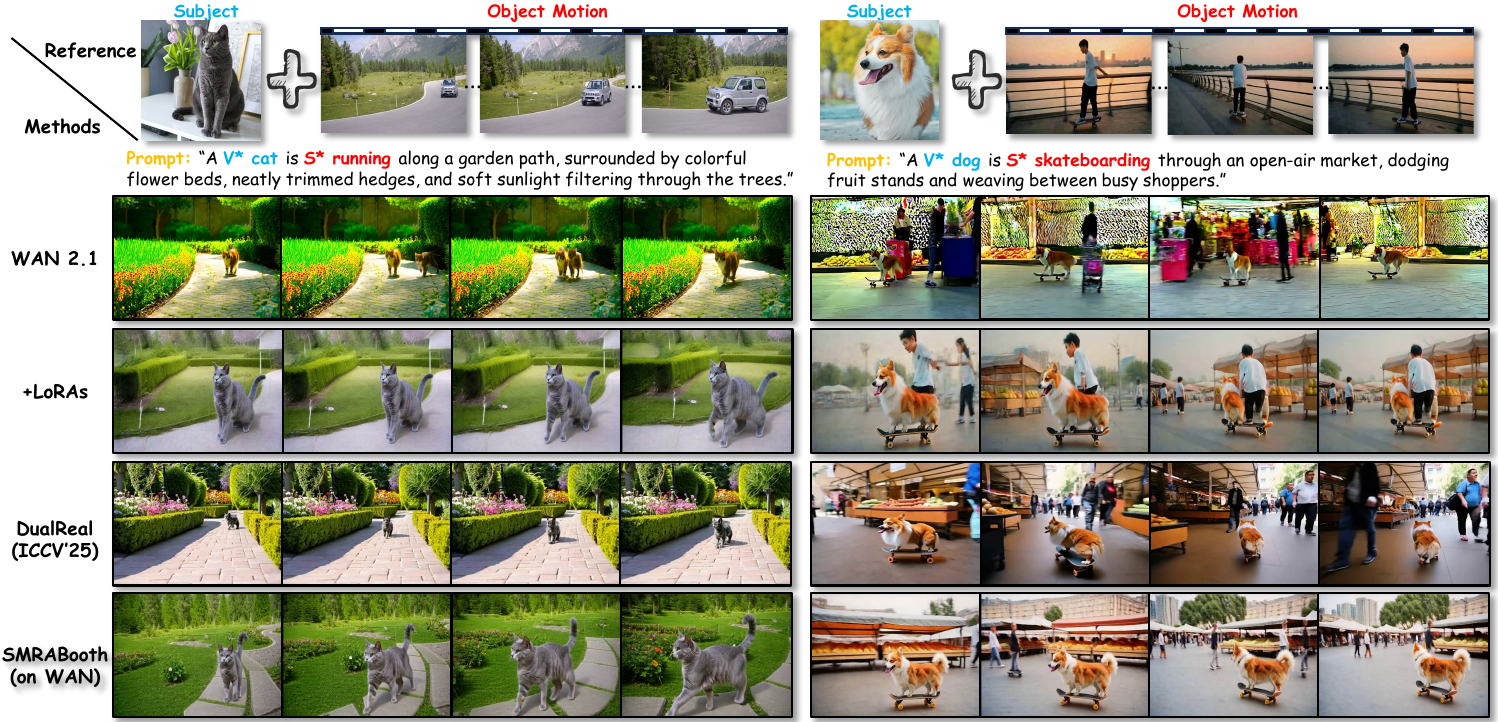}
    \vspace{-0.3cm}
    \caption{\textbf{Qualitative comparison of customized video generation on DiT-based methods.} SMRABooth preserves subject identity and motion patterns, while other methods produce artifacts and inconsistencies with the reference.}
    \label{fig_qua_joint_dit}
    \vspace{-0.6cm}
\end{figure*}

Our observation shows that LoRA is sparse in both injection locations and timing.  
Specifically, for injection locations, different layers of the WAN block specialize in either subject or motion features. 
This suggests that separating subject and motion LoRAs across layers can preserve subject and motion features separately without interference. 
However, selective layer tuning is generally less effective than full-layer fine-tuning. 
To address this, we conduct sparse experiments based on full-layer fine-tuning, evaluating subject and motion LoRAs using three metrics: CLIP-I and DINO-I for subject LoRA (measuring subject fidelity) and Motion Fidelity for motion LoRA (measuring motion similarity). 
First, we train subject and motion LoRAs individually across all layers and evaluate their inference performance without interference from the other LoRA.
To further analyze layer contributions, we traverse individual layers while setting the LoRA scale of other layers to zero.  
In Fig.~\ref{fig_timestep}(a), we normalize the metrics by assigning 100\% to full-layer fine-tuning and 0\% to the lowest performance, and visualize the changes across layers for subject LoRA and motion LoRA. 
Our results reveal that subject LoRA is primarily influenced by layers \texttt{Q, K, FFN.0}, while motion LoRA is significantly affected by layers \texttt{V, O, FFN.0, FFN.2}.  
To further validate the importance of \texttt{FFN.0} for both subject and motion LoRAs, we conduct an ablation study on Combinations\textcircled{2} and \textcircled{3} (Table~\ref{table_ablation_dit}).  

Moreover, we further analyze injection timing to decouple the influences of spatial and temporal LoRAs. Existing studies~\cite{wu2025customcrafter,li2024personalvideo} indicate that T2V models recover motion during early denoising stages and refine spatial details later. Building on this, we leverage the coarse-to-fine nature of diffusion-based generation and identify the optimal timestep $T_{point}$ through empirical analysis.  
Before $T_{point}$, subject LoRA is assigned a lower weight to prioritize motion recovery, while after $T_{point}$, it receives a higher weight to enhance subject fidelity. 
To determine $T_{point}$, we use DINO-I and CLIP-I metrics to measure frame similarity to reference images. As denoising progresses, both visual clarity and the metric scores progressively increase. By injecting both LoRAs across all timesteps, we observe trends and define $T_{point}$ as the timestep where metrics show significant improvement, signaling the onset of spatial detail refinement.  
As shown in Fig.~\ref{fig_timestep}(b), metrics increase notably between timesteps 10 and 25, indicating a refinement of spatial details. Thus, we select $T_{point}$ within this range and apply higher weights to subject LoRA after $T_{point}$ to achieve balanced subject-motion association.

\section{Experiments}\label{exp} 
\subsection{Experimental Settings}
\noindent\textbf{\emph{Implementation Details.}} For subject learning, We train the subject LoRA with a learning rate of $1.0 \times 10^{-4}$ and the rank of 32, using DINOv2-VIT-g~\cite{oquab2023dinov2} as the encoder and resizing subjects to a resolution of $512 \times 512$. We set $\lambda$ to 0.05 to balance $L_{SuRA}$ at 1/5 of $L_{region}$.
For motion learning, the motion LoRA is trained with the same learning rate, and the rank is set to 64, using SEA-RAFT~\cite{wang2024sea} as the optical flow encoder and sampling videos into 49 frames at a resolution of $576 \times 320$. We set $\alpha$ to 1.0 to achieve the best results.
During inference, we use a 50-step DDIM sampler~\cite{songdenoising} and classifier-free guidance~\cite{ho2022classifier} to generate 49-frame videos at 15 fps and $832 \times 480$ resolution. $V^*$ and $S^*$ specify subject and motion without intrinsic meanings. 
Experiments are conducted on two 96G NVIDIA H20 GPUs using the DiT~\cite{peebles2023scalable}-based WAN2.1~\cite{wan2025wan} 1.3B text-to-video model.

\noindent\textbf{\emph{Dataset.}} For subject customization, we collect 30 objects from studies like ~\cite{ruiz2023dreambooth,wumotionbooth,chen2025jointtuner}, including pets, plushies, toys, cartoons, and vehicles.  
For motion customization, we curate a dataset of 21 object motion types. Videos are sourced from datasets such as ~\cite{pont20172017,soomro2012ucf101,shi2025decouple, 11210198} and online, covering animal and vehicle movements, as well as human actions like sports and playing instruments. Some videos include the interaction between camera movement and object movement. All prompts are created and refined using GPT-4o~\cite{achiam2023gpt}.
During inference, we generate 560 combination videos by pairing subject and motion cases.
\begin{figure*}[t]
    \centering
    \includegraphics[width=0.95\textwidth]{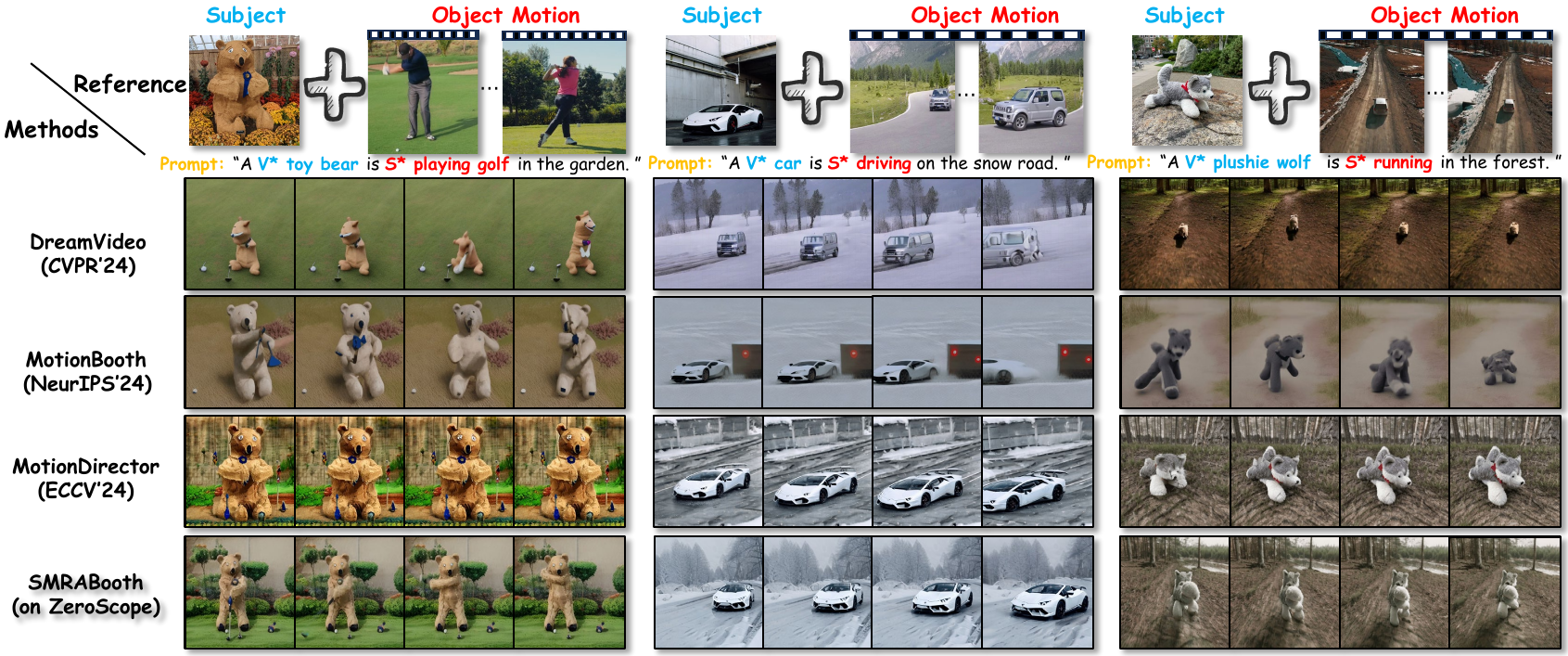}
    \vspace{-0.3cm}
    \caption{\textbf{Qualitative comparison of customized video generation on U-Net-based methods.} SMRABooth preserves subject identity and motion patterns, while other methods fail to stay faithful to the reference.}
    \label{fig_qua_joint_unet}
    \vspace{-0.3cm}
\end{figure*}
\renewcommand{\arraystretch}{0.5} 
\setlength{\tabcolsep}{2.5pt} 
\begin{table*}[t]
\centering
\caption{Quantitative experimental results for different DiT-based methods under the numerical evaluation metrics.}
\vspace{-1em}
\begin{tabular}{lccccccccc}
\toprule
\footnotesize 
\multirow{2}{*}[-1.7ex]{Method} & \multicolumn{3}{c}{\textbf{Semantic Alignment}} & \multicolumn{3}{c}{\textbf{Motion Quality}} & \multicolumn{3}{c}{\textbf{Perceptual Quality}} \\ 

\cmidrule(r){2-4} \cmidrule(r){5-7} \cmidrule(l){8-10}

& \makecell[c]{CLIP-T}\textcolor{red}{$\uparrow$} 
& \makecell[c]{CLIP-I}\textcolor{red}{$\uparrow$} 
& \makecell[c]{DINO-I}\textcolor{red}{$\uparrow$} 
& \makecell[c]{Motion\\Fidelity}\textcolor{red}{$\uparrow$} 
& \makecell[c]{Subject\\Consistency}\textcolor{red}{$\uparrow$} 
& \makecell[c]{Temporal\\Consistency}\textcolor{red}{$\uparrow$} 
& \makecell[c]{Pick\\Score}\textcolor{red}{$\uparrow$} 
& \makecell[c]{Aesthetic\\Quality}\textcolor{red}{$\uparrow$}  
& \makecell[c]{Imaging\\Quality}\textcolor{red}{$\uparrow$} 
\\ \midrule  

WAN2.1~\cite{wan2025wan}  & 0.339 & 0.586 & 0.165 & 35.18 & \textbf{95.62} & \textbf{0.988} & 19.96  & 61.09  & 65.12  \\ 
+LoRAs~\cite{hulora} & 0.314 & 0.681 & 0.464 & 60.08 & 94.33 & 0.980 & 19.85  & 56.17 & 55.90 \\
DualReal~\cite{wang2025dualreal} & 0.351 & 0.692 & 0.509 & 45.75 & 94.07 & 0.982 & 20.58 & 61.03 & 65.27\\ 
\textbf{SMRABooth (Ours)} & \textbf{0.363} & \textbf{0.700} & \textbf{0.519} & \textbf{62.89} & \underline{95.31} & \textbf{0.988} & \textbf{21.14} & \textbf{62.18} & \textbf{67.46} \\

\bottomrule
\end{tabular}
\label{table_comparison}
\vspace{-0.5cm}
\end{table*}

\noindent\textbf{\emph{Metrics.}} Following ~\cite{wei2024dreamvideo,wu2025customcrafter,huang2024vbench,zheng2025vbench,chen2025jointtuner,xu2025chain}, we evaluate our approach using nine metrics across three aspects:  
(1) \underline{Semantic Alignment}: text-video similarity (CLIP-T), image-video similarity (CLIP-I, DINO-I);  
(2) \underline{Motion Quality}: motion alignment (Motion Fidelity), subject coherence (Subject Consistency), temporal coherence (Temporal Consistency);  
(3) \underline{Perceptual Quality}: human preference (PickScore), aesthetics (Aesthetic Quality), image fidelity (Imaging Quality).

\noindent\textbf{\emph{Compared Methods.}}  
To evaluate the effectiveness of SMRABooth, we compare it with State-of-the-Art (SOTA) approaches on two backbones.  
For U-Net-based methods, including DreamVideo~\cite{wei2024dreamvideo}, MotionBooth~\cite{wumotionbooth}, and MotionDirector~\cite{zhao2024motiondirector}, we use SMRABooth (on ZeroScope~\cite{Zeroscope}). \textbf{Details are provided in the supplementary material.}  
For DiT-based methods, as DualReal~\cite{wang2025dualreal} is the latest and most representative approach, we compare original WAN2.1~\cite{wan2025wan}, WAN2.1+LoRAs~\cite{hulora}, and the DualReal~\cite{wang2025dualreal} with SMRABooth (on WAN~\cite{wan2025wan}).

\subsection{Qualitative Evaluation}
To evaluate the visual quality of synthesized videos, we compare SOTA approaches~\cite{wei2024dreamvideo, wumotionbooth, zhao2024motiondirector, wan2025wan, hulora, wang2025dualreal} with our SMRABooth. SMRABooth enables customized subject and motion generation on both DiT-based methods (Fig.~\ref{fig_qua_joint_dit}) and U-Net-based methods (Fig.~\ref{fig_qua_joint_unet}).  
In Fig.~\ref{fig_qua_joint_dit}, WAN2.1~\cite{wan2025wan} fails to preserve subject identities and motion patterns in both videos. WAN2.1+LoRAs~\cite{hulora} struggles to separate appearance information from the source video (e.g., in the 2\textsuperscript{nd} video, it mistakenly generates a man beside the dog). DualReal~\cite{zhao2024motiondirector} often fails to replicate motion patterns from the source videos, as the subject does not follow the correct motion in either case.  
In contrast, SMRABooth (on WAN) generates videos with fine-grained subject details consistent with reference images and accurately reproduces motion patterns, as shown in both videos.

In Fig.~\ref{fig_qua_joint_unet}, DreamVideo~\cite{wei2024dreamvideo} fails to preserve subject identities in the 1\textsuperscript{st}, 2\textsuperscript{nd}, and 3\textsuperscript{rd} videos. MotionBooth~\cite{wumotionbooth} struggles with motion accuracy and subject consistency (e.g., 2\textsuperscript{nd} row), while MotionDirector~\cite{zhao2024motiondirector} often fails to combine motion and subject information, producing nearly static videos (e.g., 1\textsuperscript{st} and 3\textsuperscript{rd} videos). 
In contrast, SMRABooth (on ZeroScope) generates videos that accurately retain subject details aligned with reference images and effectively reproduce motion patterns, as shown in all videos.

\renewcommand{\arraystretch}{0.9} 
\begin{table}[t]
\centering
\caption{Quantitative User Studies on DiT-based Methods.}
\vspace{-10pt}
\footnotesize 
\begin{tabular}{l|c|c|c|c}\toprule
    Method         
    & \makecell[c]{Prompt\\Alignment}
    & \makecell[c]{Motion\\Similarity} 
    & \makecell[c]{Appearance\\Similarity}
    & \makecell[c]{Video\\Quality}
    \\ \midrule  
    WAN2.1~\cite{wan2025wan} & 3.688$\pm$0.051 &1.764$\pm$0.040 & 1.498$\pm$0.022 & 3.855$\pm$0.047  \\
    +LoRAs~\cite{hulora} & 3.523$\pm$0.060 &3.185$\pm$0.058 & 3.444$\pm$0.058 & 3.242$\pm$0.062  \\
    DualReal~\cite{wang2025dualreal} & 3.919$\pm$0.063 &3.019$\pm$0.055 & 3.527$\pm$0.051 & 3.968$\pm$0.049  \\
    \midrule
    \textbf{SMRABooth} & \textbf{4.228$\pm$0.041} & \textbf{3.468$\pm$0.060} & \textbf{4.178$\pm$0.044} & \textbf{4.244$\pm$0.040} \\  
    \bottomrule 
\end{tabular}
\vspace{-0.7cm}
\label{table_userstudy}
\end{table}
\renewcommand{\arraystretch}{0.4} 
\setlength{\tabcolsep}{2.5pt} 
\begin{table*}[t]
\centering
\caption{Quantitative ablation studies on each component.}
\vspace{-1em}
\begin{tabular}{lccccccccc}
\toprule

\multirow{2}{*}[-1.7ex]{Method} & \multicolumn{3}{c}{\textbf{Semantic Alignment}} & \multicolumn{3}{c}{\textbf{Motion Quality}} & \multicolumn{3}{c}{\textbf{Perceptual Quality}} \\ 

\cmidrule(r){2-4} \cmidrule(r){5-7} \cmidrule(l){8-10}

& \makecell[c]{CLIP-T}\textcolor{red}{$\uparrow$} 
& \makecell[c]{CLIP-I}\textcolor{red}{$\uparrow$} 
& \makecell[c]{DINO-I}\textcolor{red}{$\uparrow$} 
& \makecell[c]{Motion\\Fidelity}\textcolor{red}{$\uparrow$} 
& \makecell[c]{Subject\\Consistency}\textcolor{red}{$\uparrow$} 
& \makecell[c]{Temporal\\Consistency}\textcolor{red}{$\uparrow$} 
& \makecell[c]{Pick\\Score}\textcolor{red}{$\uparrow$} 
& \makecell[c]{Aesthetic\\Quality}\textcolor{red}{$\uparrow$}  
& \makecell[c]{Imaging\\Quality}\textcolor{red}{$\uparrow$} 
\\ \midrule  

w/o $l_{SuRA}$  & 0.338 & 0.667 & 0.467  & 62.15 &95.29 &0.987 &21.07  & 60.17  & 67.15  \\ 
w/o $l_{MoRA}$ & 0.338 & 0.686 & 0.501 & 60.02 & 94.30 & 0.984 & 20.89  & 59.55 & 61.29 \\
\midrule
Combination\textcircled{1} & 0.355 & 0.652 & 0.460 & 61.77 & 94.90 & 0.982 & 20.40 & 59.76 & 66.29\\ 
Combination\textcircled{2} & 0.350 & 0.649 & 0.357 & 57.80 & \textbf{95.89} & \underline{0.989} & 20.45 & 62.05 & 65.75\\ 
Combination\textcircled{3} & 0.358 & 0.684 & 0.480 & 56.12 & 94.54 & \textbf{0.991} & 20.90 & 61.52 & 67.30\\  
\midrule
\textbf{SMRABooth (Ours)} & \textbf{0.363} & \textbf{0.700} & \textbf{0.519} & \textbf{62.89} & \underline{95.31} & 0.988 & \textbf{21.14} & \textbf{62.18} & \textbf{67.46} \\

\bottomrule
\end{tabular}
\label{table_ablation_dit}
\vspace{-0.3cm}
\end{table*}
\subsection{Quantitative Evaluation}  
We evaluate the performance of SMRABooth(on WAN) against several SOTA methods~\cite{wan2025wan, hulora, wang2025dualreal} on multiple subject-video-prompt pairs.  
Additional results for U-Net-based comparisons, including ~\cite{wei2024dreamvideo, wumotionbooth, zhao2024motiondirector}, \textbf{are provided in the supplementary material}.

\noindent\textbf{\emph{Objective Evaluation.}}  
As shown in Table~\ref{table_comparison}, SMRABooth outperforms SOTA methods across all three evaluation categories: Semantic Alignment, Motion Quality, and Perceptual Quality. Specifically:  
(1) SMRABooth achieves the highest scores in text-video alignment (CLIP-T: 0.363), visual similarity with reference images (CLIP-I: 0.700, DINO-I: 0.519), and motion fidelity (62.89), surpassing all baselines. This demonstrates its ability to faithfully preserve the reference subject, video, and prompt.  
(2) SMRABooth achieves a Motion Fidelity score of 62.89, significantly outperforming DualReal (45.75), highlighting its superior capability in learning and replicating motion patterns.  
(3) SMRABooth also obtains the highest scores in PickScore (21.14), Aesthetic Quality (62.18), and Imaging Quality (67.46), showcasing its capacity to generate videos that are not only visually appealing but also of exceptional quality.
Overall, SMRABooth establishes SOTA in text alignment, motion quality, and perceptual fidelity, validating its effectiveness for customized video generation.

\noindent\textbf{\emph{User Studies.}}  
We conducted user studies to evaluate the effectiveness of SMRABooth. A total of 100 participants voted on 84 pairs of customized videos, which were randomly placed to reduce bias and evaluated based on four criteria: prompt alignment, motion similarity, appearance similarity, and video quality. Each video was rated on a scale of 1 to 5, resulting in 8,400 ratings.
As shown in Table~\ref{table_userstudy}, SMRABooth achieved the highest scores across all criteria compared to other SOTA methods.  
We also performed an analysis on the $95\%$ confidence interval and found that our SMRABooth outperformed baseline methods with statistically significant results.

\begin{figure}[t]
    \vspace{-0.2cm}
    \centering
    \includegraphics[width=0.47\textwidth]{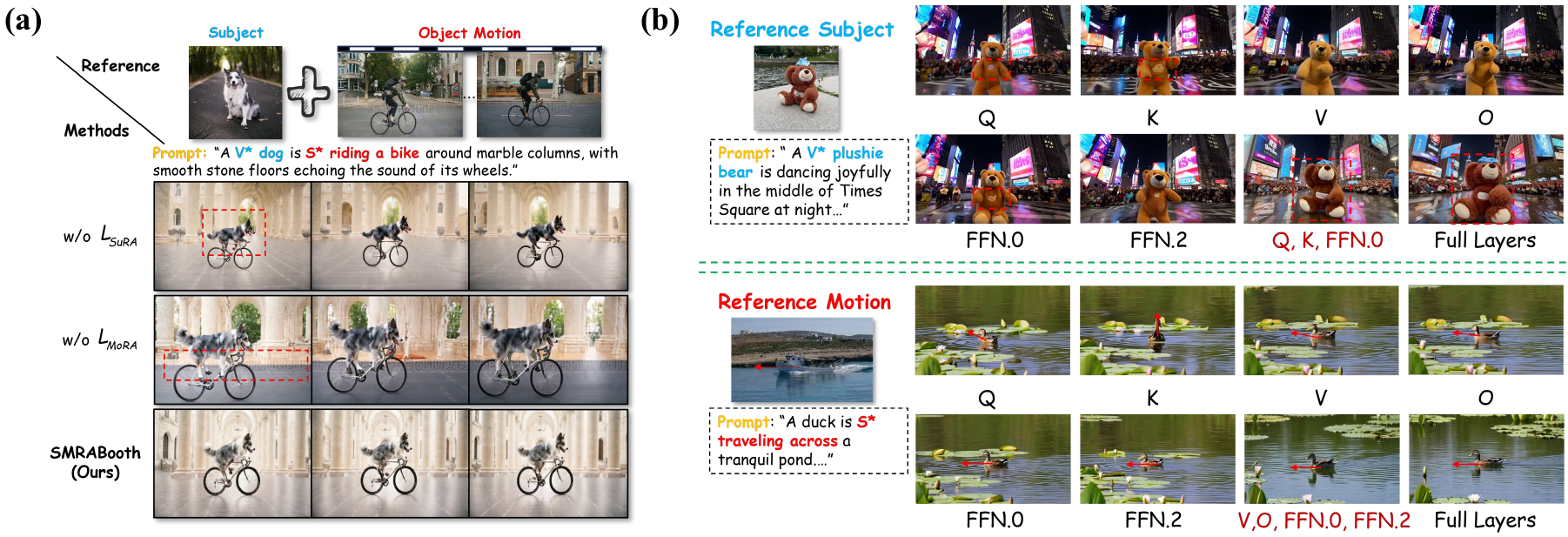}
    \vspace{-10pt}
    \caption{Qualitative comparison for the ablation study. (a) shows the ablation results for $l_{SuRA}$ and $l_{MoRA}$, and (b) visualizes the sparsely selected layers compared to the full-layer LoRAs.}
    \label{fig_ablation}
    \vspace{-0.6cm}
\end{figure}
\subsection{Ablation Study}
To assess the contribution of each component in SMRABooth, we conduct comprehensive ablation studies focusing on the core modules of SMRABooth and the impact of our Subject-Motion Association Injection strategy, as well as its integration with different LoRA layer combinations.

\noindent\textbf{Effect of Subject Representation Alignment module.}
As shown in Table~\ref{table_ablation_dit} and Fig.~\ref{fig_ablation}(a), incorporating $l_{SuRA}$ significantly enhances the preservation of global structure and semantic consistency. This module introduces high-level spatial information, enabling the model to maintain fidelity to reference identities while reducing reliance on low-level details. Consequently, the results show notable improvements in semantic alignment metrics CLIP-I and DINO-I.

\noindent\textbf{Effect of Motion Representation Alignment module.}
The $l_{MoRA}$ effectively captures object-level motion information, enabling the generation of structurally coherent motion trajectories. As shown in Table~\ref{table_ablation_dit} and Fig.~\ref{fig_ablation}(a), $l_{MoRA}$ improves Motion Fidelity and Temporal Consistency, resulting in realistic and consistent motion patterns while successfully disentangling appearance from motion dynamics.

\noindent\textbf{Effect of Subject-Motion Association Decoupling.}
We evaluate three different combinations of LoRA layers to analyze their effects on the model's performance (as shown in Table.~\ref{table_ablation_dit}) and Fig.~\ref{fig_ablation}(b):
(1) \textbf{Combination\textcircled{1}:} Full layers for both subject and motion LoRAs. This configuration significantly reduces overall video quality due to the decoupling of subject and motion LoRAs, which leads to low-quality video generation.
(2) \textbf{Combination\textcircled{2}:} \texttt{Q, K} for subject LoRA and \texttt{V, O, FFN.0, FFN.2} for motion LoRA. Here, we observe a degradation in CLIP-I and DINO-I due to the absence of FFN.0 in the subject LoRA, which negatively impacts subject learning (as analyzed in Sec.~\ref{sub:Inject}).
(3) \textbf{Combination\textcircled{3}:} \texttt{Q, K, FFN.0} for subject LoRA and \texttt{V, O, FFN.2} for motion LoRA. This setup significantly reduces Motion Fidelity, as the absence of FFN.0 in the motion LoRA adversely impacts motion learning (as discussed in Sec.~\ref{sub:Inject}).
(4) \textbf{SMRABooth (Ours):} \texttt{Q, K, FFN.0} are selected for subject LoRA, and \texttt{V, O, FFN.0, FFN.2} for motion LoRA. By focusing on the most impactful layers for each, this configuration achieves optimal video quality while preserving reference subject and motion patterns. As shown in Fig.~\ref{fig_ablation}(b), our sparse LoRAs achieve a similar effect to full-layer fine-tuning in both subject and motion.

As shown in the \textbf{supplementary material}, we evaluate different $T_{point}$ choices, applying lower subject LoRA weights before $T_{point}$ and higher weights afterward.  
For small $T_{point}$, the video appears nearly static due to subject LoRA interfering with motion LoRA.  
For large $T_{point}$ values, the video loses consistency with reference images as subject LoRA is applied too late.  
We set $T_{point} = 15$ and double the subject LoRA weight after $T_{point}$.
\section{Conclusion}\label{con}  
We present SMRABooth, a novel framework that introduces global representation alignment with stage-by-stage customization for DiT-based architectures. SMRABooth ensures global semantic consistency and structural fidelity through subject alignment, while motion alignment maintains temporal coherence and realistic motion. The subject-motion association decoupling strategy further balances appearance fidelity and video coherence.  
Extensive experiments show that SMRABooth maintains diffusion model flexibility while consistently generating high-quality, text-aligned outputs matching reference subjects and videos on both DiT-based and U-Net-based methods.
In future work, we aim to expand SMRABooth to handle more complex and diverse customization tasks, broadening its applicability across various scenarios.

{
    \small
    \bibliographystyle{ieeenat_fullname}
    \bibliography{main}
}

\setcounter{page}{1}
\maketitlesupplementary
\section{Overview}\label{overview}
\makeatother
The supplementary material includes the following sections:  
\begin{enumerate}
    \item \textcolor{black}{Additional implementation details.}
    \item \textcolor{black}{A detailed discussion of our methods.}
    \item \textcolor{black}{Additional qualitative results generated by our DiT-based method.}
    \item \textcolor{black}{Extended details for our U-Net-based method.}
    \item \textcolor{black}{More qualitative results generated by our U-Net-based method.}
    \item \textcolor{black}{A local anonymous project page and a demo video.}
    \item \textcolor{black}{A folder containing the videos generated by our method.}
\end{enumerate}

\section{More implementation details}\label{Implementation}
\noindent\subsection{Further explanation of the ablation experiments for our DiT-Based method}\label{sub:abexp}  
In Sec. 4.4, we propose a subject-motion association decoupling injection strategy that sparsifies LoRAs in injection timing. Specifically, we adjust subject LoRA weights at a critical timestep $T_{point}$ to balance appearance fidelity and video coherence. Lower subject LoRA weights are applied before $T_{point}$ to prioritize motion generation, while higher weights are applied afterward to enhance subject identity preservation and temporal consistency. Here, we conduct two ablation studies to evaluate the choice of $T_{point}$ and the subject LoRA scale before $T_{point}$.  

\begin{figure}[t]
    \centering
    \includegraphics[width=0.5\textwidth]{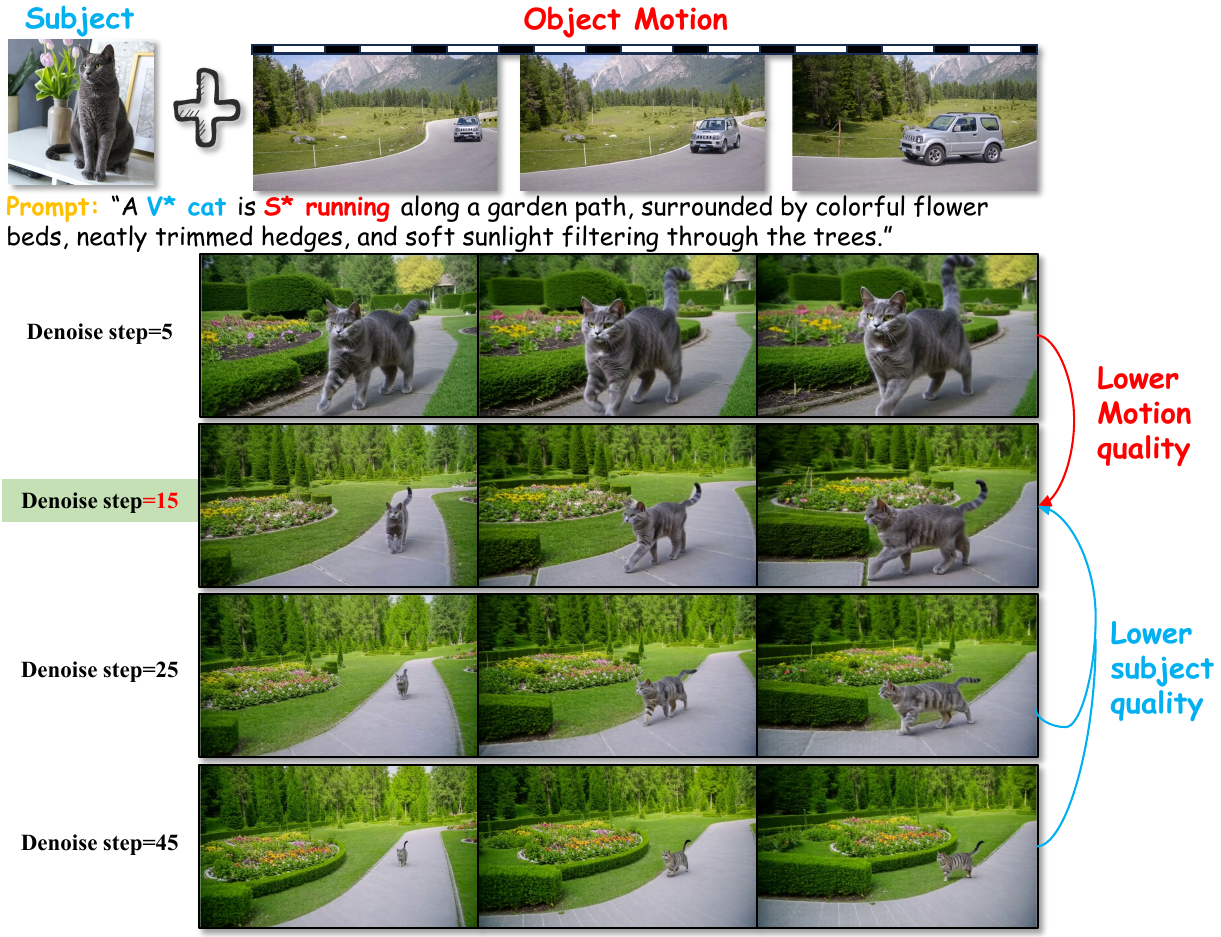}
    \vspace{-20pt}
    \caption{\textbf{Ablation study on injection timing $T_{point}$}: We analyze four choices: 5, 15, 25, 45 and find that applying lower subject LoRA before denoise step 15, as described in Sec. 3.4, achieves the best performance.}
    \label{fig_ablation_timestep}
    \vspace{-0.2cm}
\end{figure}

\noindent\textbf{Effect of Sparse LoRA Injection Timing.}  
To determine $T_{point}$, we conduct an ablation study alongside the theoretical analysis in Sec. 4.4, as shown in Fig.~\ref{fig_ablation_timestep}. Setting $T_{point}$ too early causes subject LoRA to interfere with motion LoRA, disrupting motion generation (e.g., Fig.~\ref{fig_ablation_timestep}, Denoise step=5), as the model focuses on motion in the early denoising stages.  
Conversely, setting $T_{point}$ too late results in subjects inconsistent with the reference (e.g., Fig.~\ref{fig_ablation_timestep}, Denoise step=25 and Denoise step=45). At this stage, the model emphasizes fine-grained details, but a low subject LoRA scale allows the text prior to dominate, causing deviations from the reference.  
Based on our analysis, we set $T_{point}$ to 15 (see Fig.~\ref{fig_ablation_timestep}, Denoise step=15), striking a balance between maintaining subject appearance and preserving temporal dynamics for coherent motion and accurate subject representation.
\begin{figure}[t]
    \centering
    \includegraphics[width=0.5\textwidth]{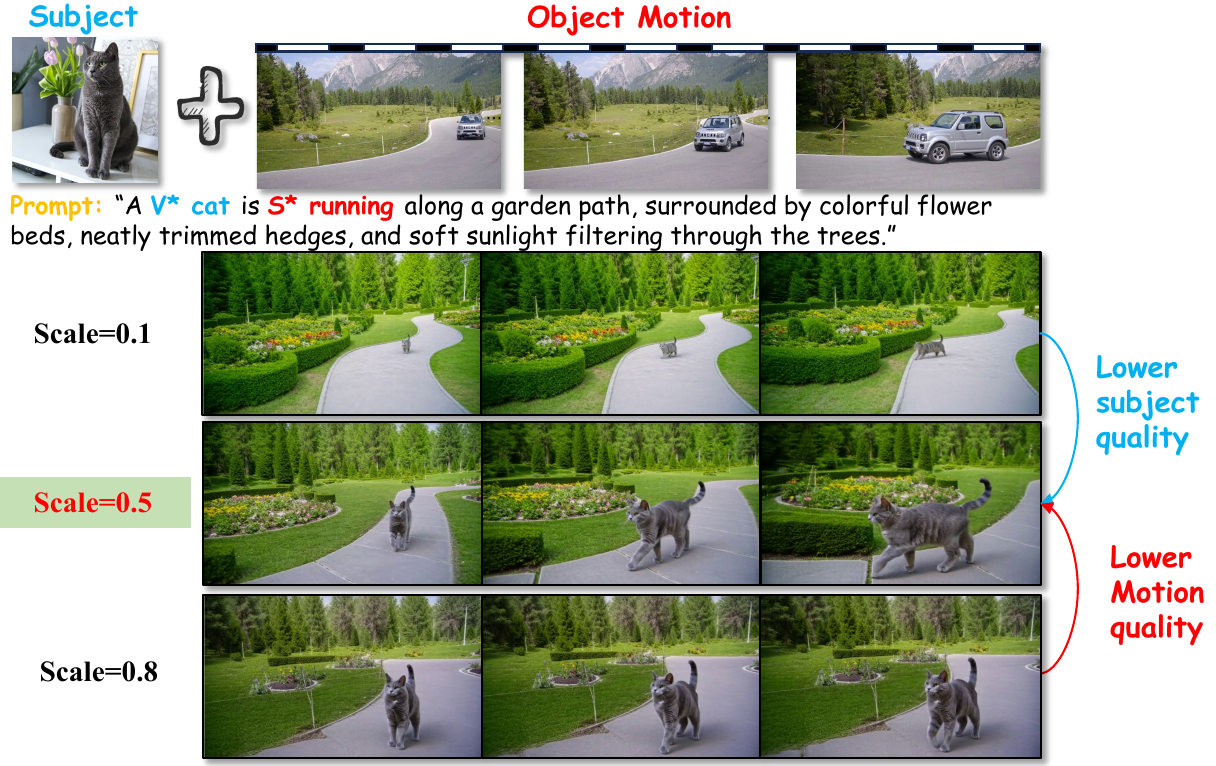}
    \caption{\textbf{Ablation study on subject LoRA scale before $T_{point}$}: By exploring different scales, we find that a subject LoRA scale of 0.5 achieves the best performance before $T_{point}$.}
    \label{fig_ablation_scale}
    \vspace{-0.1cm}
\end{figure}

\noindent\textbf{Comparison of Different Scales of subject LoRA Injection.}  
In our subject-motion association decoupling injection strategy, we adjust subject LoRA weights at $T_{point}$ to balance appearance fidelity and video coherence. Lower subject LoRA weights are applied before $T_{point}$, while higher weights are applied afterward to achieve a smooth trade-off between subject identity preservation and temporal consistency.  
To analyze the effect of the subject LoRA scale, we conduct an ablation study (see Fig.~\ref{fig_ablation_scale}). Setting the subject LoRA scale too small (e.g., Fig.~\ref{fig_ablation_scale}, Scale=0.1) results in poor preservation of the subject's shape and structure, leading to inaccurate subject generation and loss of critical information. On the other hand, setting the scale too large (e.g., Fig.~\ref{fig_ablation_scale}, Scale=0.8) interferes with the motion LoRA, making the video appear static and losing dynamic motion consistency.  
Based on our analysis, we set the subject LoRA scale to 0.5 (see Fig.~\ref{fig_ablation_scale}, Scale=0.5), achieving the best balance between subject appearance and temporal dynamics for coherent motion and accurate subject representation.

\noindent\subsection{Details for Temporal Representation Alignment}\label{sub:ITRA}
In Eq.9, we present an equation to reverse $z_t$ to $z_0$ with one-step denoising.  
A more comprehensive explanation of the derivation of Eq.9 is provided in Alg.~\ref{alg:flow_matching}.
\begin{algorithm}[htb]
    \small
    \caption{Prediction Function for $z_0$ in Flow Matching}
    \label{alg:flow_matching} 
    \setlength{\baselineskip}{15pt} 
    \begin{algorithmic}[1]
        \State \textbf{Prediction function for $z_0$ in Flow Matching}
        \State In Flow Matching: $z_t = (1-t) \cdot z_0 + t \cdot z_1$
        \State \quad $\Rightarrow z_0 = \frac{z_t - t \cdot z_1}{1-t}$ \quad \textit{...(1)}
        \State where $z_1$ is noise, model predicts the velocity: $v_\theta \approx z_1 - z_0$
        \State \quad $\Rightarrow z_1 \approx v_\theta + z_0$ \quad \textit{...(2)}
        \State \textbf{Substitute (2) into (1):}
        \State $z_0 = \frac{z_t - t \cdot (v_\theta + z_0)}{1-t}$
        \State $z_0 = \frac{z_t - t \cdot v_\theta - t \cdot z_0}{1-t}$
        \State $z_0 \cdot (1-t) = z_t - t \cdot v_\theta - t \cdot z_0$
        \State $z_0 = z_t - t \cdot v_\theta$ \quad \textit{...(3)}
        \State  where in WAN, $v_\theta =  u(z_t, c_{txt}, t; \theta)$ \quad \textit{...(4)}
        \State \textbf{Substitute (4) into (3):}
        \State \textbf{Therefore:} $z_0 = z_t - t \cdot u(z_t, c_{txt}, t; \theta)$
    \end{algorithmic}
\end{algorithm}

Additionally, Eq.8 requires reversing the latents and using a 3D VAE to decode them on CUDA, which results in significant computational overhead. To address this issue, we adopt a sliding window technique to efficiently select a subset of latents for decoding. 

The sliding window is set to a size of 6 in latent space and moves by 2 frames at a time. This means that at each step, the 3D VAE processes $1 + 4 \times (6 - 1) = 21$ frames simultaneously. Considering the temporal dependency characteristics of the 3D VAE, we discard the first 5 frames and retain the subsequent 16 frames. During experiments, we confirm that the preserved 16 frames effectively retain motion patterns similar to those of the source video frames at corresponding positions, making them suitable for our motion representation.

\begin{figure}[t]
    \centering
    \includegraphics[width=0.5\textwidth]{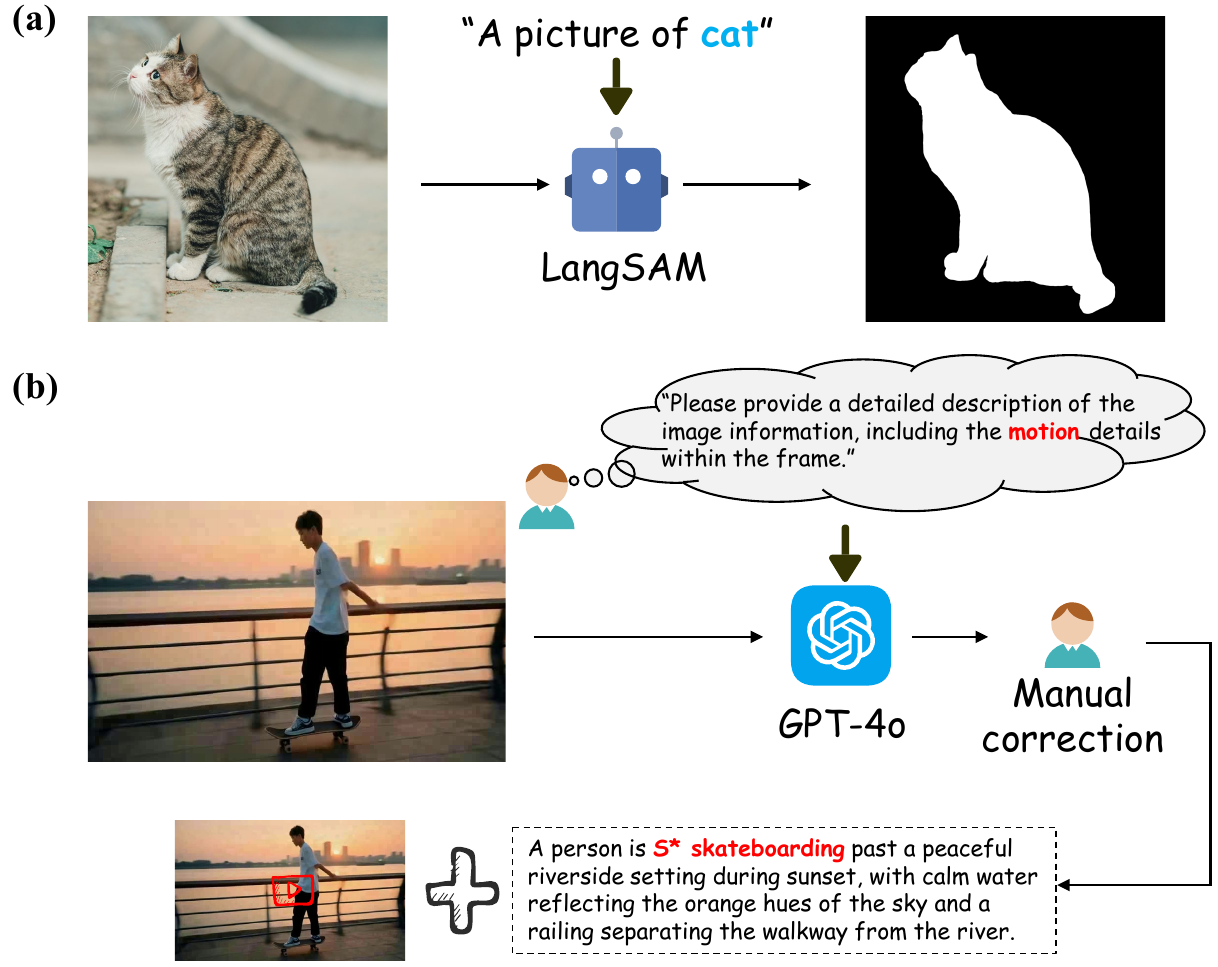}
    \caption{Illustration of our dataset construction. (a) shows the processing details for subject images. We guide the LangSAM model with text prompts to segment the subject area, resulting in a binarized mask. (b) shows the processing details for motion videos. We sample a frame from the videos, use GPT-4o to generate a detailed caption for the frame, manually refine the prompt, add special tokens, and pair the caption with the source video.}
    \label{fig_dataset}
    \vspace{-0.3cm}
\end{figure}
\noindent\subsection{Dataset Construction Details}
The pipeline for constructing our training dataset is illustrated in Fig.~\ref{fig_dataset}. For the subject images, we use LangSAM to generate binarized masks. First, we guide the segmentation model LangSAM by injecting accurate descriptions of the target object to ensure precise segmentation of the subject area. LangSAM can process images of any resolution. After segmentation, we annotate each subject image with the caption "A picture of V* $<$subject name$>$" and combine it with the original image and the generated mask to create the training data. 
For the motion videos, we first sample one frame from the source video and use GPT-4o to generate a caption that provides a detailed description of the visual content and its motion characteristics. After obtaining the caption, we manually refine it by removing redundant information, such as audio-related descriptions. Finally, the corrected caption is paired with the source video to construct the motion dataset.
During inference, we use GPT-4o to combine subjects and motions in pairs and generate diverse and creative background descriptions for them. We exclude illogical combinations, such as "a car playing the piano" or "a car playing basketball," to ensure realistic scenarios.

\noindent\subsection{Baseline Details and Training Cost}  
For DualReal, we adopt CogVideoX-5B as the text-to-video backbone. We run 1,000 training steps for each test case, and each output contains 49 frames at a resolution of $720 \times 480$ pixels.  
For WAN2.1+LoRAs fine-tuning, we follow the official WAN2.1 1.3B training code. LoRA uses a learning rate of $1 \times 10^{-4}$ with 300 subject LoRA steps and 400 motion LoRA steps, the same as SMRABooth. Each output contains 49 frames at a resolution of $480 \times 832$ pixels.  
For WAN2.1 1.3B, we test the backbone's native reasoning capabilities. Each output contains 49 frames at a resolution of $480 \times 832$ pixels.

For the training cost, our SMRABooth requires approximately 30 minutes to train a single subject LoRA and about one hour to train a single motion LoRA. In contrast, DualReal requires joint training of two hours for the combined LoRAs of subject and motion.
\noindent\subsection{Detailed introduction of metrics}
We establish a comprehensive evaluation framework across three dimensions: Semantic Alignment, Motion Quality and Perceptual Quality, using nine metrics.
\begin{itemize}
\item \textcolor{black}{\textbf{Semantic Alignment.}} 
(1) \textbf{CLIP-T:} This metric evaluates the alignment between text prompts and generated videos by calculating the frame-wise cosine similarity between their embeddings, derived from the CLIP~\cite{radford2021learning} model.  
(2) \textbf{CLIP-I:} This metric assesses the visual-semantic correspondence by comparing the embeddings of reference images and generated video frames. These embeddings are obtained using the image encoder from CLIP~\cite{radford2021learning}.  
(3) \textbf{DINO-I:} Similar to CLIP-I, this metric measures the visual-semantic correspondence; however, it utilizes the DINO~\cite{oquab2023dinov2} vision transformer encoder to compute and compare embeddings of reference images and generated frames.
For these three metrics, we evaluate them on our generated video frame by frame and compute their final scores by averaging the results across all frames.
\item \textcolor{black}{\textbf{Motion Quality.}}  
(1) \textbf{Motion Fidelity:} Evaluates the consistency of motion patterns by leveraging CoTracker3~\cite{karaev2024cotracker}, a model designed for diffusion-motion-transfer~\cite{yatim2024space}.  
(2) \textbf{Subject Consistency:} Assesses whether the appearance of the subject (e.g., characters) remains consistent across different frames in the video, as implemented in VBench~\cite{huang2024vbench}.  
(3) \textbf{Temporal Consistency:} Quantifies frame-to-frame consistency by calculating the average cosine similarity of CLIP~\cite{radford2021learning} image embeddings across all frame pairs in the generated video.  
\item \textcolor{black}{\textbf{Perceptual Quality.}}  
(1) \textbf{PickScore:} Predicts human preference scores using PickScore~\cite{kirstain2023pick}, with results averaged at the frame level.  
(2) \textbf{Aesthetic Quality:} Measures artistic merit using the LAION aesthetic predictor, implemented via VBench~\cite{huang2024vbench}.  
(3) \textbf{Imaging Quality:} Evaluates distortions in generated frames, such as overexposure, noise, and blurriness, as assessed via VBench~\cite{huang2024vbench}.
\end{itemize}
\begin{figure}[t]
    \centering
    \includegraphics[width=0.5\textwidth]{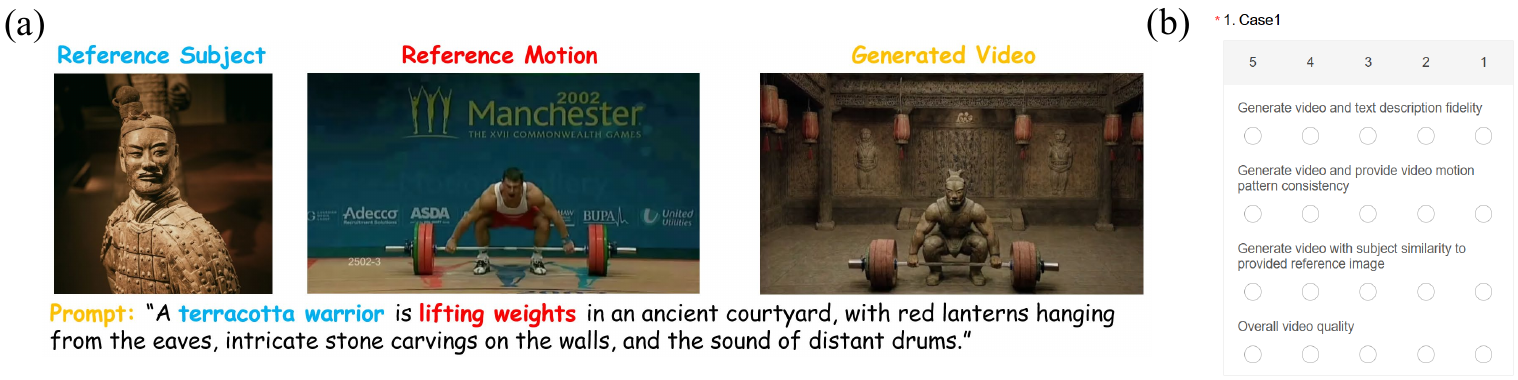}
    \vspace{-10pt}
    \caption{\textbf{Human evaluation questionnaire format.} (a) presents the reference subject image, reference video, and the customized video generated by the model. Participants are then asked to complete the evaluation form in (b), rating the quality of the generated video based on Prompt Alignment, Motion Similarity, Appearance Similarity and Video Quality.}
    \label{fig_human}
    \vspace{-0.5cm}
\end{figure}

\noindent\subsection{User study interface}\label{sub:USI}
During the user study, we provide each case video generated by WAN2.1, WAN2.1+LoRAs, DualReal, and our SMRABooth (WAN) for evaluation based on four questions. Each question is rated on a scale from 1 to 5 for the following criteria:  
(1) The accuracy of generating the video to match the text descriptions (\textit{Prompt Alignment}).  
(2) Consistency between the generated video and the provided motion mode (\textit{Motion Similarity}).  
(3) The similarity between the main body of the generated video and the reference image provided (\textit{Appearance Similarity}).  
(4) The overall quality of the video (\textit{Video Quality}).
Fig.~\ref{fig_human} shows the format of our questionnaire format.
\section{Discussion}\label{discussion}
\noindent\textbf{Methods Discussion.}  
While existing methods like VideoJAM~\cite{chefer2025videojam} leverage optical flow to improve motion quality for T2V models, our Motion Representation Alignment differs significantly:  
(1) \textbf{Task Perspective:} VideoJAM focuses on enhancing the backbone model's motion quality, aiming for generalizable motion across diverse prompts and scenarios. In contrast, SMRABooth specializes in learning fixed motion patterns from reference videos. 
Additionally, SMRABooth leverages optical flow primarily for motion feature extraction decoupled from appearance, further reducing the coupling with appearance features through sparse LoRAs.
(2) \textbf{Method Perspective:} VideoJAM relies on extensive training with video and optical flow, whereas SMRABooth adopts a lightweight fine-tuning framework. VideoJAM utilizes in-context learning, which requires a large amount of training data, while SMRABooth's temporal representation alignment enables motion pattern learning from a single video.

\noindent\textbf{Limitations.} 
While SMRABooth excels in customized subject and motion generation, it has some limitations.  
One key limitation is its inability to handle multi-object customized generation, a related and important area left unexplored in this work. Expanding SMRABooth in this direction could greatly enhance its versatility for more complex scenarios.  
Another challenge is the lack of open-source datasets for subject and video customization, limiting our ability to conduct more extensive experiments to further validate the model's performance. Developing or accessing such datasets is crucial for advancing research in this field.
\section{Additional qualitative results generated by our DiT-based method.}\label{sec:add_dit}
\clearpage
\begin{figure*}[t]
    \centering
    \includegraphics[width=1.0\textwidth]{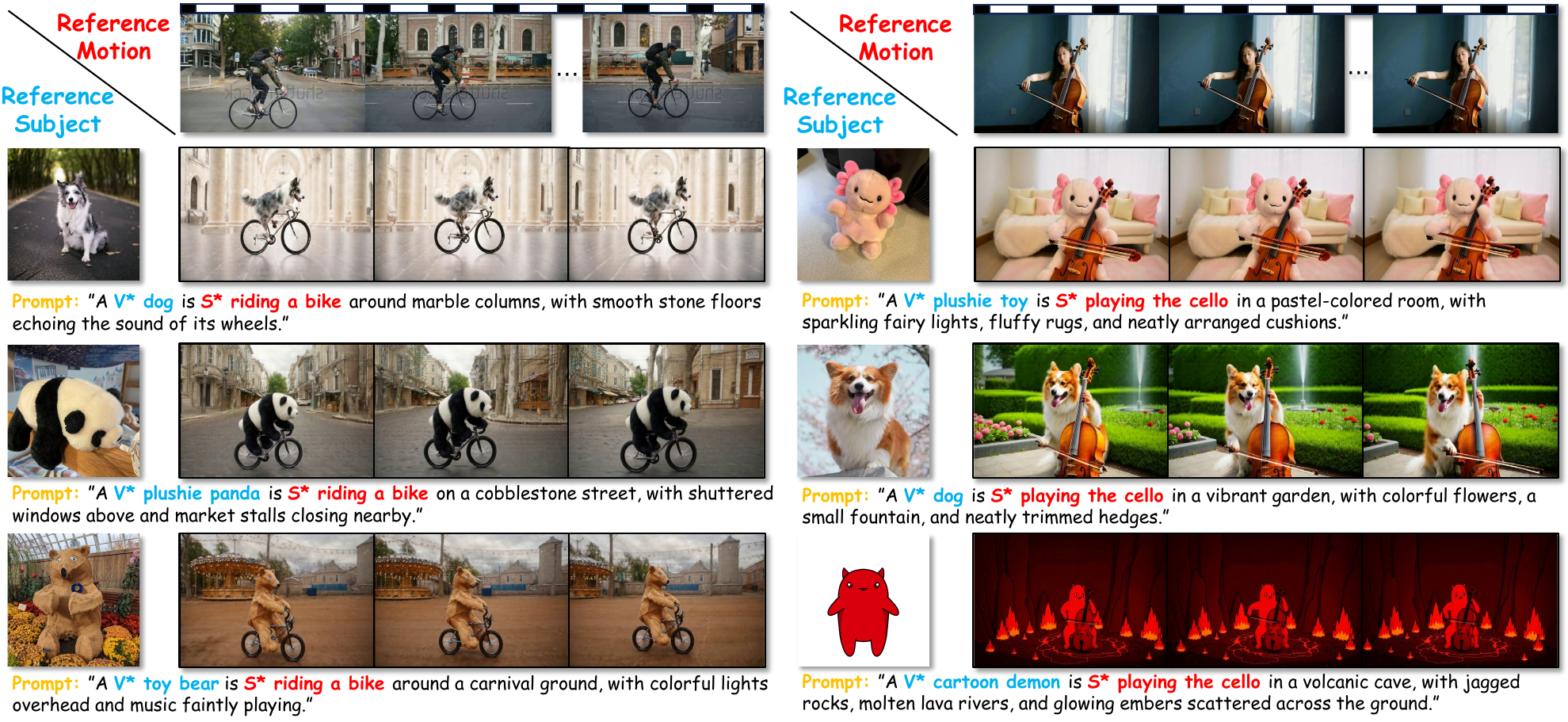}
    \vspace{-10pt}
    \caption{\textbf{More qualitative results of our joint customization for subject and motion.} In this case, we use a set of videos to guide our model in learning the motion concept. SMRABooth generates customized videos that accurately preserve subject appearance and motion patterns while remaining faithful to text prompts.}
    \label{fig_joint_1}
   \vspace{-1cm}
\end{figure*}
\begin{figure*}[t]
    \centering
    \includegraphics[width=1.0\textwidth]{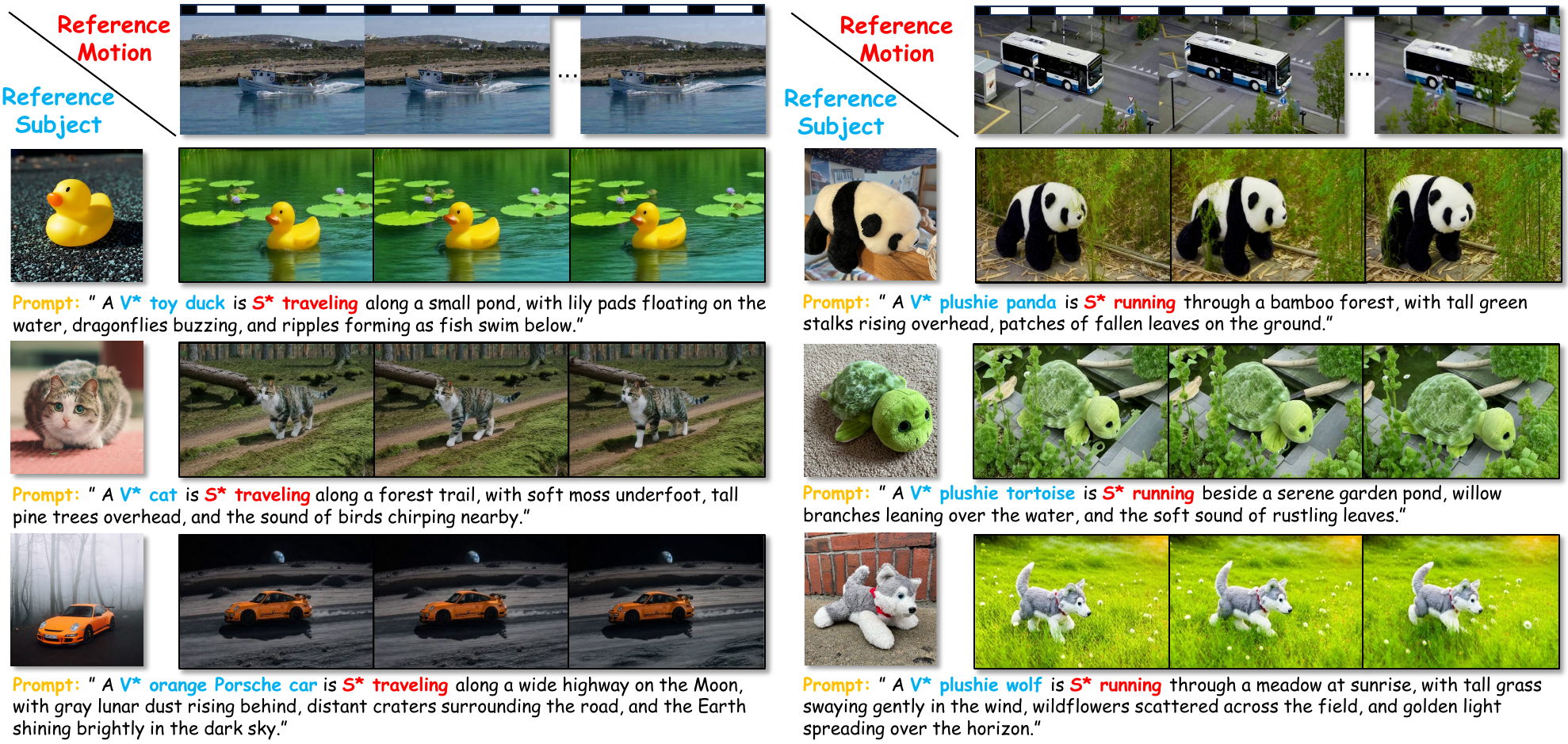}
    \vspace{-10pt}
    \caption{\textbf{More qualitative results of our joint customization for subject and motion.} In this case, we use a set of videos to guide our model in learning the motion concept. SMRABooth generates customized videos that accurately preserve subject appearance and motion patterns while remaining faithful to text prompts.}
    \label{fig_joint_2}
   \vspace{-1cm}
\end{figure*}
\begin{figure*}[t]
    \centering
    \includegraphics[width=1.0\textwidth]{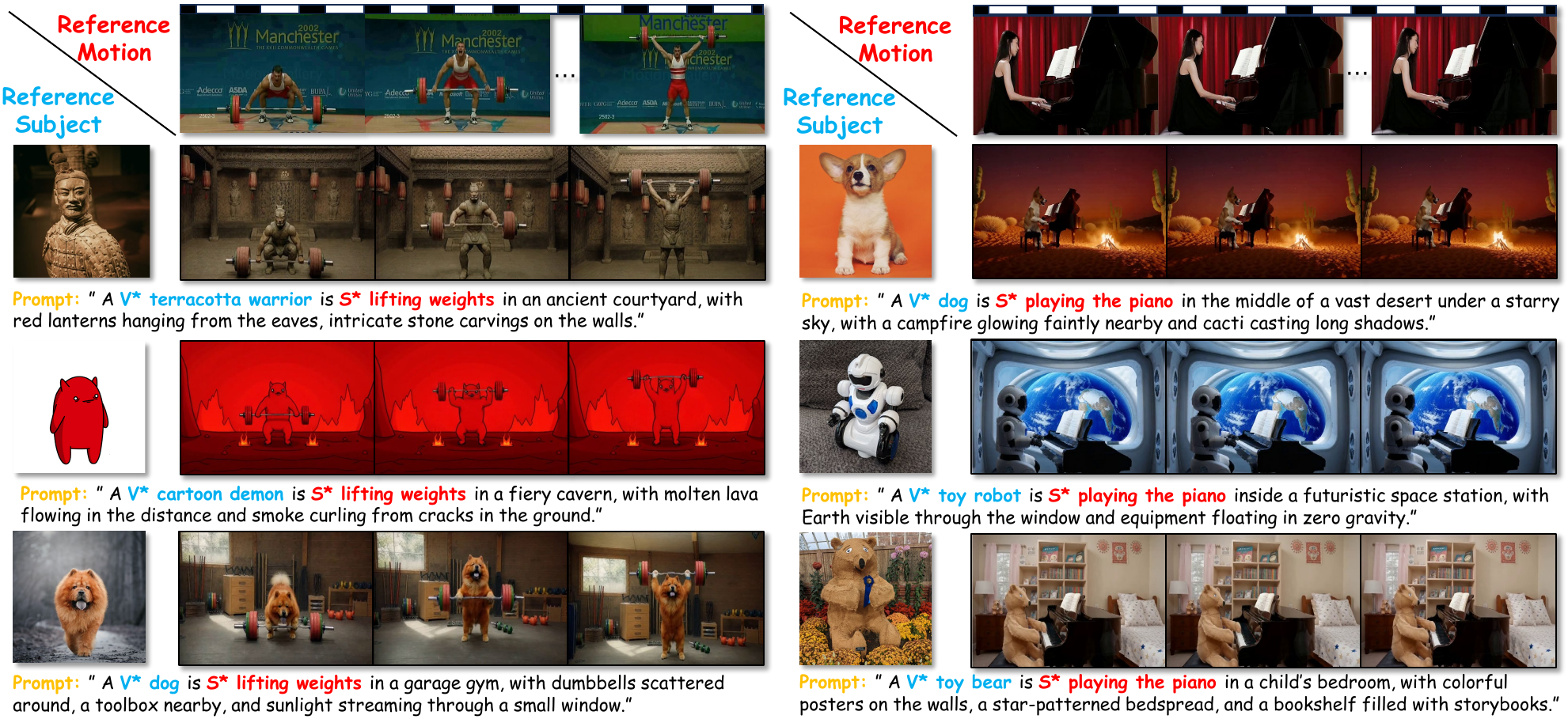}
    \vspace{-10pt}
    \caption{\textbf{More qualitative results of our joint customization for subject and motion.} In this case, we use a set of videos to guide our model in learning the motion concept. SMRABooth generates customized videos that accurately preserve subject appearance and motion patterns while remaining faithful to text prompts.}
    \label{fig_joint_3}
   \vspace{-0.5cm}
\end{figure*}
\begin{figure*}[t]
    \centering
    \includegraphics[width=1.0\textwidth]{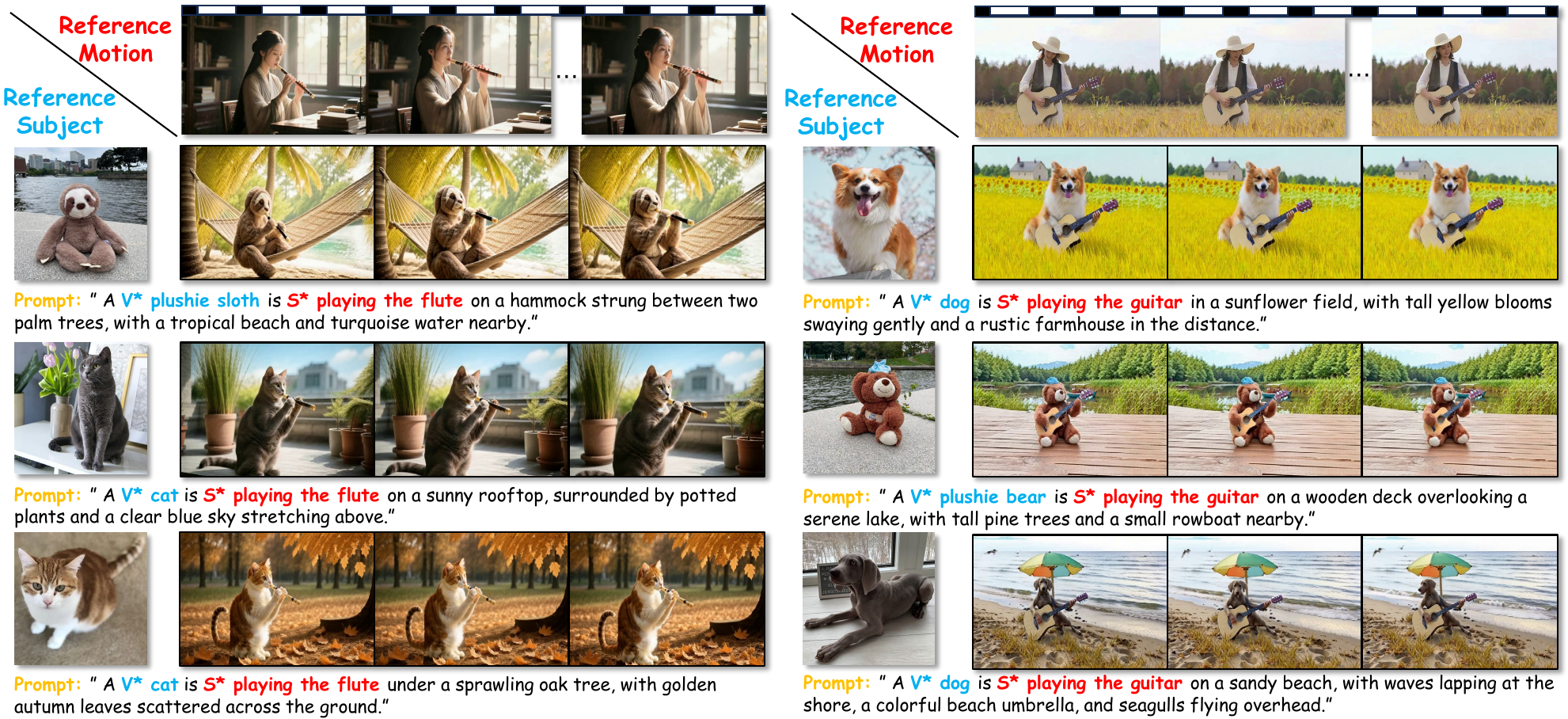}
    \vspace{-10pt}
    \caption{\textbf{More qualitative results of our joint customization for subject and motion.} In this case, we use a set of videos to guide our model in learning the motion concept. SMRABooth generates customized videos that accurately preserve subject appearance and motion patterns while remaining faithful to text prompts.}
    \label{fig_joint_4}
   \vspace{-0.5cm}
\end{figure*}
\begin{figure*}[t]
    \centering
    \includegraphics[width=1.0\textwidth]{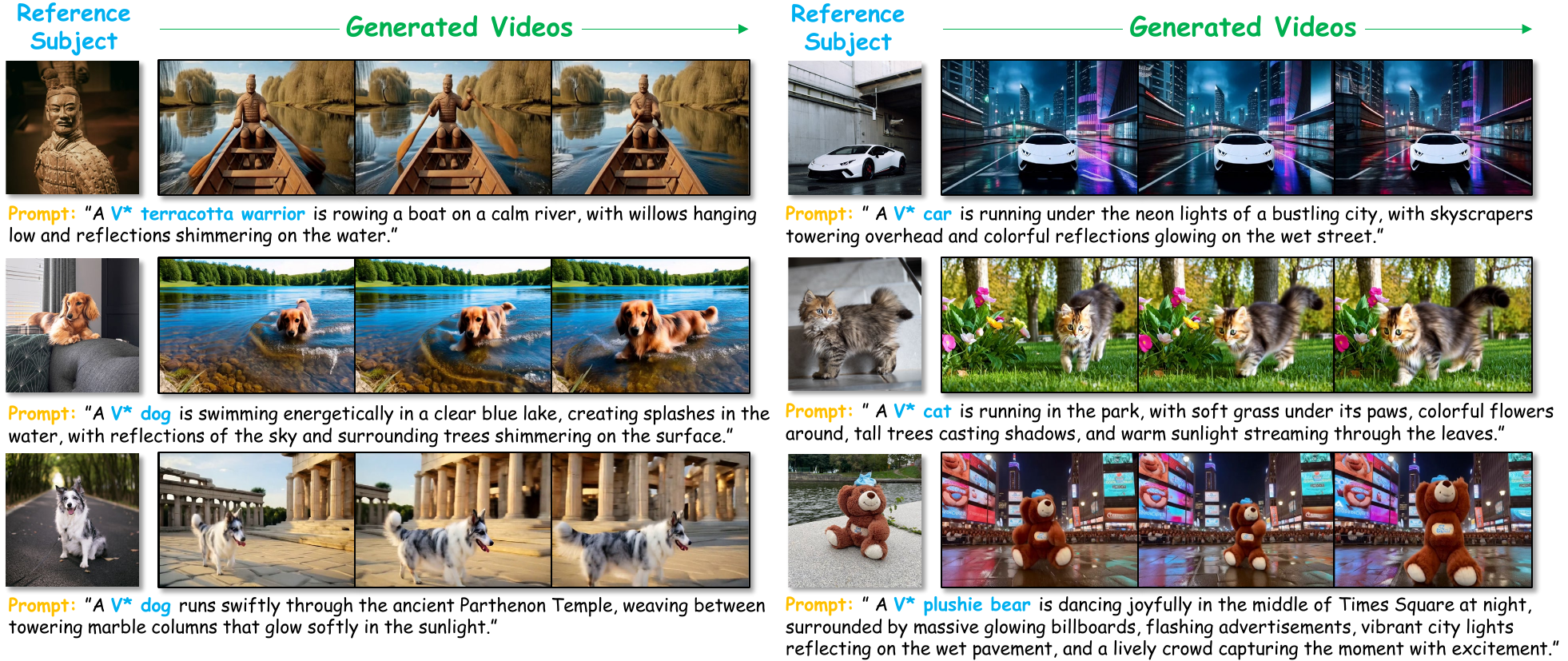}
    \vspace{-10pt}
    \caption{\textbf{More qualitative results of our customization for subject.} In this case, We have customized and generated a variety of different subjects, including various world historical sites and natural landscapes. Our cases fully demonstrate the accurate extraction of theme features and the strong generalization capabilities of our model.}
    \vspace{-0.2cm}
    \label{fig_onlysubject}
\end{figure*}
\begin{figure*}[t]
    \centering
    \includegraphics[width=1.0\textwidth]{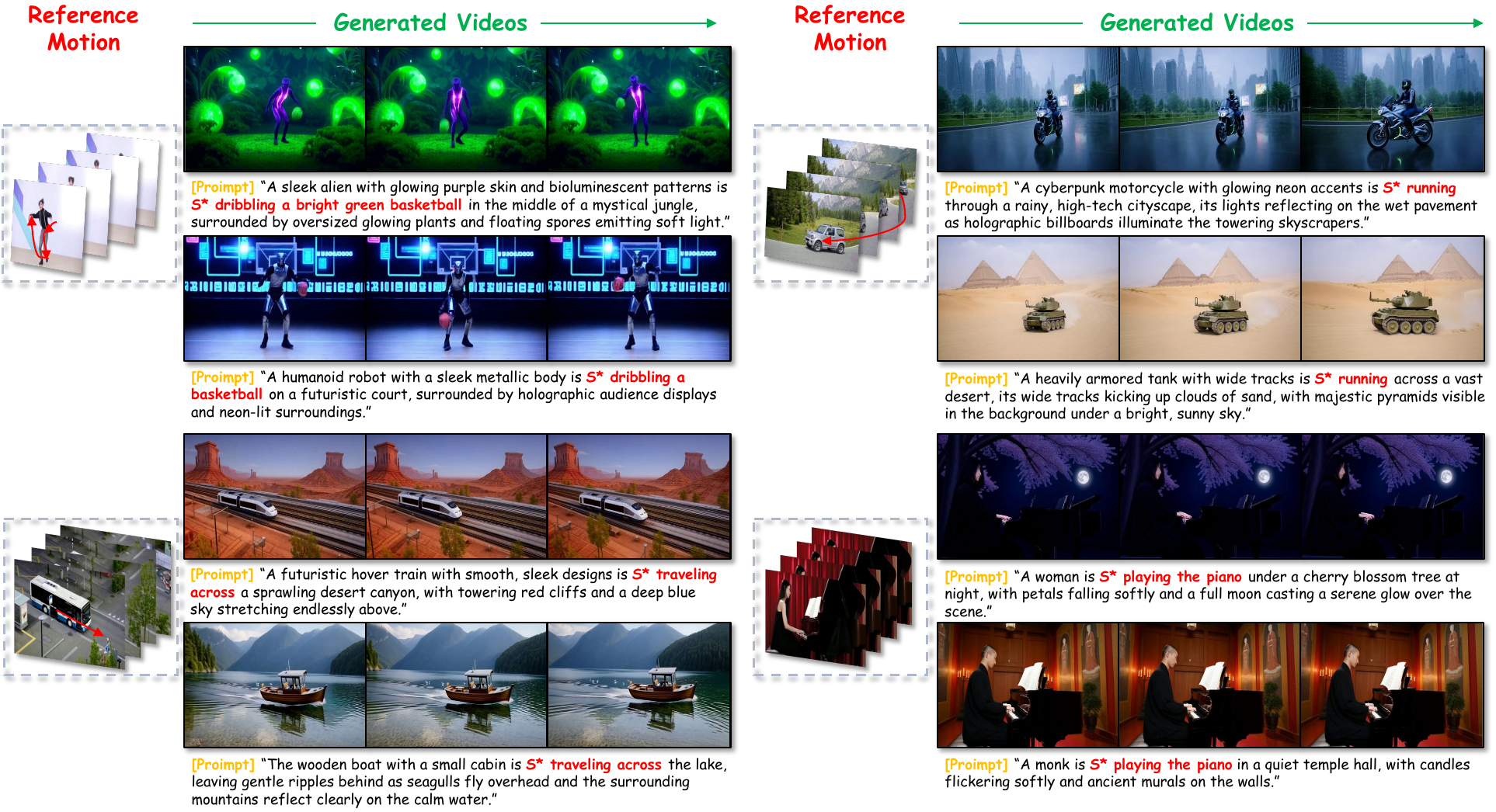}
    \vspace{-10pt}
    \caption{\textbf{More qualitative results of our customization for motion.} In this case, We have customized and generated a variety of different motion. Our cases fully demonstrate the accurate extraction of theme features and the strong generalization capabilities of our model.}
    \label{fig_onlymotion}
    \vspace{-0.5cm}
\end{figure*}

\clearpage
\begin{figure}[htbp]
    \centering
    \includegraphics[width=0.45\textwidth]{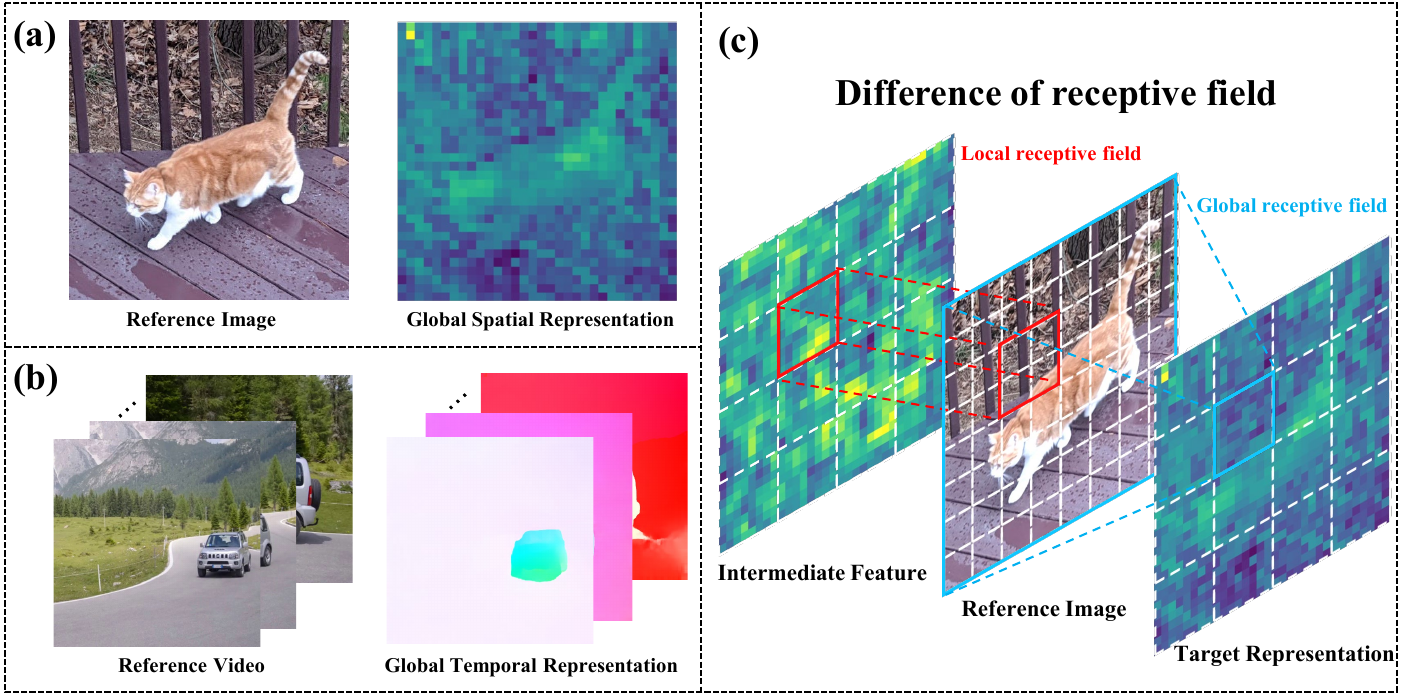}
    \caption{(a) Global spatial representations capture the global spatial structure and semantic information from the reference image.  (b) Global temporal representations capture object-level motion trajectories and motion trends from the reference video.  (c) Visualization of the receptive field of a patch in the final feature map. The target patch captures global features, while intermediate features focus only on local features.}
    \vspace{-5pt}
    \label{fig_receptivefield}
\end{figure}
\begin{figure}[htbp]
    \centering
    \includegraphics[width=0.45\textwidth]{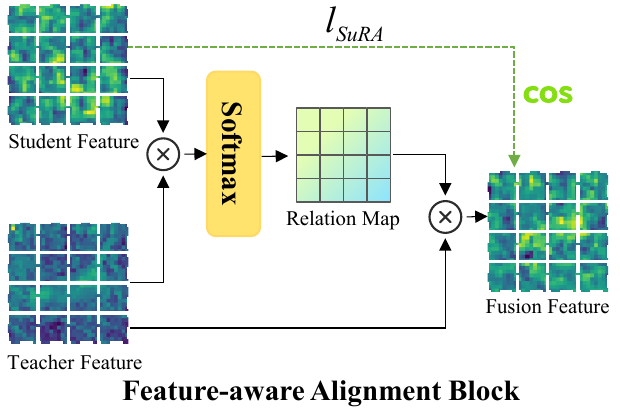}
    \caption{The design of the RAA module: It features a structure similar to cross-attention but effectively bridges the gap between receptive fields of varying sizes.}
    \vspace{-5pt}
    \label{fig_FAA}
\end{figure}
\begin{figure*}[t]  
    \centering
    \includegraphics[width=1.0\textwidth]{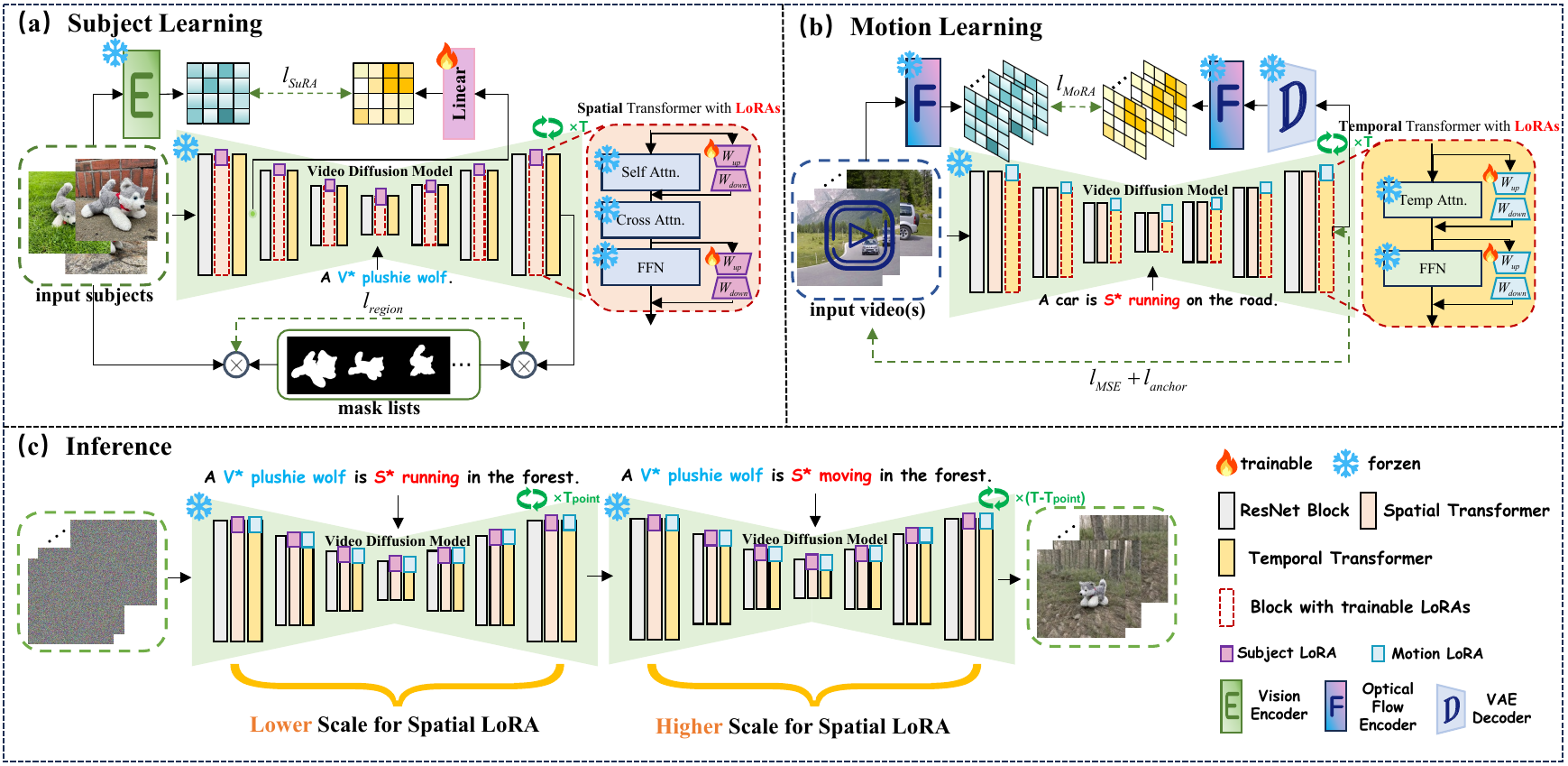}
    \vspace{-0.5cm}
    \caption{\textbf{Overview of SMRABooth for \textcolor{red}{U-Net-based SMRABooth}.} The framework divides customized video generation into two stages: subject learning and motion learning. }
    \label{fig_pipeline_zeroscope}
\end{figure*}
\section{Details for our U-Net-based method}\label{sec:U-Net}
\begin{table*}[htbp]
\centering
\caption{Quantitative experimental results for different methods under the numerical evaluation metrics.}
\vspace{-1em}
\setlength{\tabcolsep}{4pt} 
\renewcommand{\arraystretch}{0.8} 
\begin{tabular}{lcccccccc}
\toprule

\multirow{2}{*}[-1.7ex]{Method} & \multicolumn{4}{c}{\textbf{Objective evaluation}} & \multicolumn{4}{c}{\textbf{User study}} \\ 

\cmidrule(r){2-5} \cmidrule(){6-9} 

& \makecell[c]{CLIP-T}\textcolor{red}{$\uparrow$} & \makecell[c]{CLIP-I}\textcolor{red}{$\uparrow$} & \makecell[c]{DINO-I}\textcolor{red}{$\uparrow$} & \makecell[c]{T. Cons.}\textcolor{red}{$\uparrow$}  
& \makecell[c]{Prompt\\Alignment} & \makecell[c]{Motion\\Similarity} & \makecell[c]{Appearance\\Similarity} & \makecell[c]{Video\\Quality} \\ \midrule  

DreamVideo  & 0.298 & 0.609 & 0.302 & 0.970 & 2.751 & 2.867 & 2.676 & 2.767 \\ 
MotionBooth & 0.301 & 0.689 & 0.449 & 0.954 & 2.890 &2.819  & 2.915 & 2.870 \\ 
MotionDirector & 0.295 & 0.759 & 0.597 & \textbf{0.989} & 3.023 & 2.915 &3.165 &3.078\\ 
\textbf{SMRABooth(Ours)} & \textbf{0.329}   & \textbf{0.760}  & \textbf{0.612}  & \underline{0.986} & \textbf{3.488} & \textbf{3.501} & \textbf{3.543} & \textbf{3.499} \\ 

\bottomrule
\end{tabular}
\label{table_comparison_zeroscope}
\vspace{-0.3cm}
\end{table*}
\noindent\subsection{Technical details}\label{sub:tec}
For U-Net-based methods, the main challenge lies in the mismatch of receptive fields between U-Net and ViT. This inconsistency in receptive fields between encoder representations and U-Net features can cause semantic information confusion during direct alignment.
Thus, we introduce the \textbf{R}epresentation-\textbf{A}ware \textbf{A}lignment (\textbf{RAA}) block (Fig.~\ref{fig_receptivefield}(c) and Fig.~\ref{fig_FAA}).  
This block integrates each patch of local features with the entirety of global representations, significantly augmenting the ability of local features to aware global spatial structure and high-level semantic information.  
Using these settings to train a spatial low-rank matrix (LoRA) within spatial transformers, we can accurately preserve the subject's appearance from reference images.  
As shown in Fig.~\ref{fig_receptivefield}(c), the ViT-based target network, with its larger receptive fields and more feature channels, captures richer semantic context per pixel compared to the U-Net-based network, which is limited by smaller receptive fields. 
Pixel-wise alignment of representations can lead to suboptimal generation quality due to mismatches in receptive field size and semantic richness. 
While using a homogeneous encoder seems like an intuitive solution, experiments reveal it is less effective than a heterogeneous encoder, as detailed in the supplementary materials.
To address this issue, we propose a \textbf{R}epresentation-\textbf{A}ware \textbf{A}lignment (RAA) block, which first fuses the feature distributions of the two architectures instead of directly computing the loss between them. 
This approach helps bridge the representational gap between the ViT-based vision encoder and the U-Net-based.
We first compute the relation map between the intermediate and target features and fuse the two features patch by patch. The relation map is defined as:
\begin{equation}\label{eqn-14}
R = \text{softmax}\left(\frac{h_{\phi}(z_t) \cdot \mathbf{y^*}^{\top}}{\sqrt{d}}\right),
\end{equation}
The fusion feature $\mathbf{x}^*$ is then calculated as:
\begin{equation}\label{eqn-15}
\mathbf{x^*} = R  \cdot \mathbf{y}^*,
\end{equation}
Finally, feature alignment is realized through the loss function, defined as:
\begin{equation}\label{eqn-16}
\mathcal{L}_{\mathrm{SuRA}}(\theta) = -\mathbb{E}_{\mathbf{x}_{*},\boldsymbol{\epsilon},t}\left[\frac{1}{N}\sum_{n=1}^{N}sim(\mathbf{y}^{*[n]}, \mathbf{x}^{*[n]})\right],
\end{equation}
where $n$ is the patch index, and $sim(\cdot, \cdot)$ is a pre-defined similarity function.

Moreover, to prevent subject LoRA from overfitting to the background of the subject image during training, we introduce masks $M$ generated by SAM~\cite{kirillov2023segment}, which enforce the model to focus only on the subject region. Formally, the masked MSE loss is defined as:
\begin{equation}\label{eqn-17}
\mathcal{L}_{\mathrm{region}} = \mathbb{E}_{\mathbf{z},\epsilon\sim\mathcal{N}(\mathbf{0},\mathbf{I}),t,\mathbf{c}}\left[||(\epsilon - \epsilon_\theta(\mathbf{z}_t, \mathbf{c}_i, t)) \cdot \mathbf{M}||_2^2\right],
\end{equation}
where $M$ represents the mask applied to the subject region. 
During training, the overall loss function is defined as:
\begin{equation}\label{eqn-18}
\mathcal{L} = \mathcal{L}_\mathrm{region} + \lambda \mathcal{L}_{\mathrm{SuRA}},
\end{equation}
where $\lambda > 0$ is a hyperparameter that balances the subject representation alignment loss.
\subsection{Experiment evaluation}
\noindent\textbf{\emph{Quantitative Evaluation.}}  
As shown in Table~\ref{table_comparison_zeroscope}, SMRABooth outperforms SOTA methods in text-video alignment, visual similarity to reference images, and temporal consistency. Specifically:  
(1) Compared to DreamVideo~\cite{wei2024dreamvideo}, SMRABooth improves CLIP-T from 0.298 to 0.329, CLIP-I from 0.609 to 0.760, DINO-I from 0.302 to 0.612, and Temporal Consistency from 0.970 to 0.986.  
(2) Compared to MotionBooth~\cite{wumotionbooth}, SMRABooth raises CLIP-T from 0.301 to 0.329, CLIP-I from 0.689 to 0.760, and Temporal Consistency from 0.954 to 0.986.  
(3) Compared to MotionDirector~\cite{wang2024motionctrl}, SMRABooth improves CLIP-T from 0.295 to 0.329 while maintaining CLIP-I at 0.760. MotionDirector’s higher Temporal Consistency is due to still-image-like outputs lacking realistic motion.  
In summary, SMRABooth excels in text alignment, temporal coherence, and fidelity to reference images and videos.
\renewcommand{\arraystretch}{0.8} 
\begin{table}[t] 
\centering
\caption{Quantitative ablation studies on each component in a subset of our training set. We select 15 subject-motion pairs.}
\vspace{-10pt}
\begin{tabular}{l|c|c|c|c}\toprule
    Method         & CLIP-T\textcolor{red}{$\uparrow$} & CLIP-I\textcolor{red}{$\uparrow$}
    & DINO-I\textcolor{red}{$\uparrow$} & T. Cons.\textcolor{red}{$\uparrow$} \\ \midrule
    w/o $l_{SuRA}$ &0.343   &0.742   &0.523 &0.982  \\
    w/o $RAA$ &0.338   &0.744   &0.561 &0.985  \\
    w/o $l_{MoRA}$  &0.323  &0.693   &0.489 &0.987  \\
    \midrule
    \textbf{Ours}  & \textbf{0.345} & \textbf{0.754} & \textbf{0.594} & \textbf{0.988} \\ \bottomrule 
\end{tabular}
\label{table_ablation_zeroscope}
\vspace{-0.3cm}
\end{table}

\noindent\textbf{\emph{Effect of $l_{SuRA}$.}}  
As shown in Table~\ref{table_ablation_zeroscope}, $l_{SuRA}$ provides high-level spatial information, helping the model preserve global structure and semantic consistency. This ensures fidelity to reference identities while avoiding over-reliance on low-level details, resulting in more coherent outputs.  

\noindent\textbf{\emph{Effect of $RAA$.}}  
As shown in Table~\ref{table_ablation_zeroscope}, directly using pre-trained visual encoder features as alignment targets degrades the model’s features, leading to suboptimal results. The $RAA$ block resolves this by fusing heterogeneous features before alignment, allowing the model to better utilize high-level semantic information and improve subject generation quality.  

\noindent\textbf{\emph{Effect of $l_{MoRA}$.}}  
As shown in Table~\ref{table_ablation_zeroscope}, $l_{MoRA}$ provides object-level motion information, enabling the model to capture global motion trends and maintain coherent motion trajectories. This ensures consistent and realistic motion while disentangling appearance from motion dynamics.

\section{Additional qualitative results generated by our U-Net-based method.}\label{sec:add_UNet}
\begin{figure*}[t]
    \centering
    \includegraphics[width=0.95\textwidth]{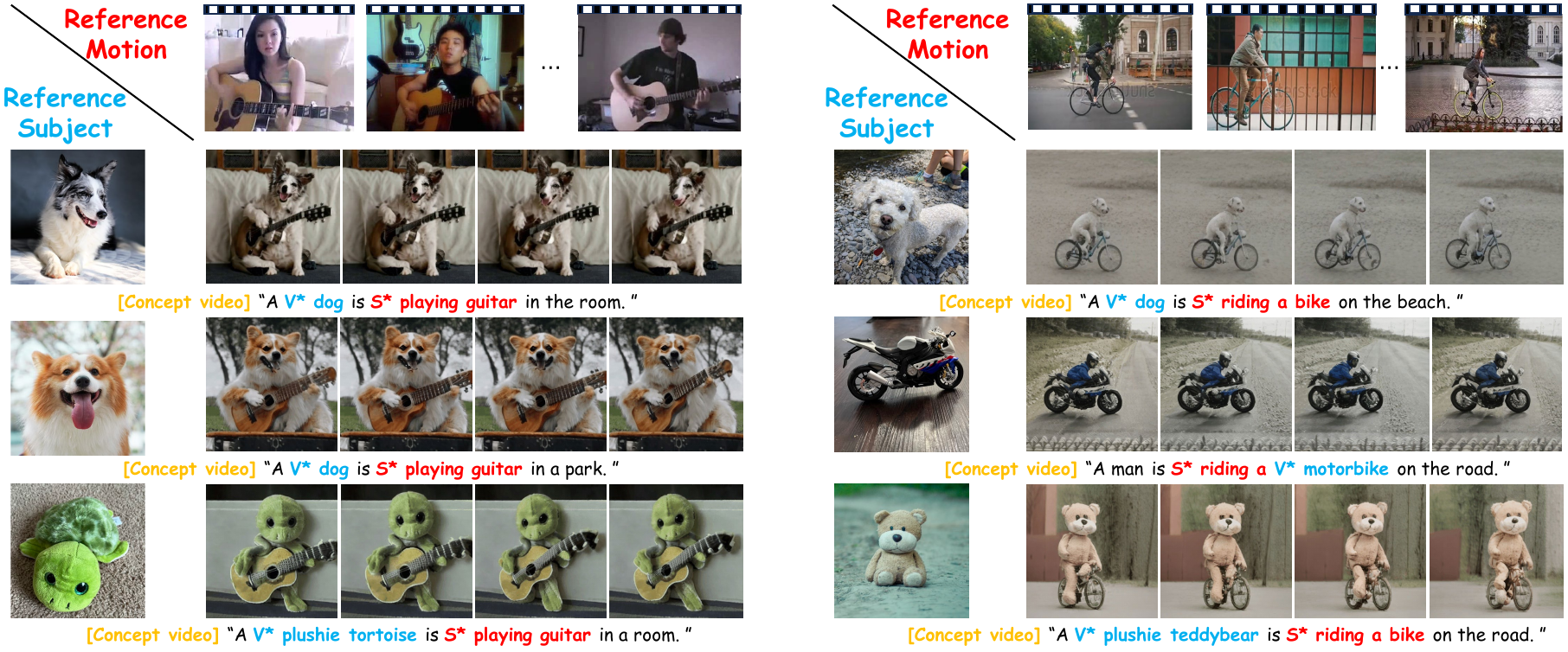}
     \vspace{-5pt}
    \caption{\textbf{More qualitative results of our joint customization for subject and motion.} In this case, we use a set of videos to guide our model in learning the motion concept. SMRABooth generates customized videos that accurately preserve subject appearance and motion patterns while remaining faithful to text prompts.}
    \label{figconcept1}
\end{figure*}
\begin{figure*}[t]
    \centering
    \includegraphics[width=0.95\textwidth]{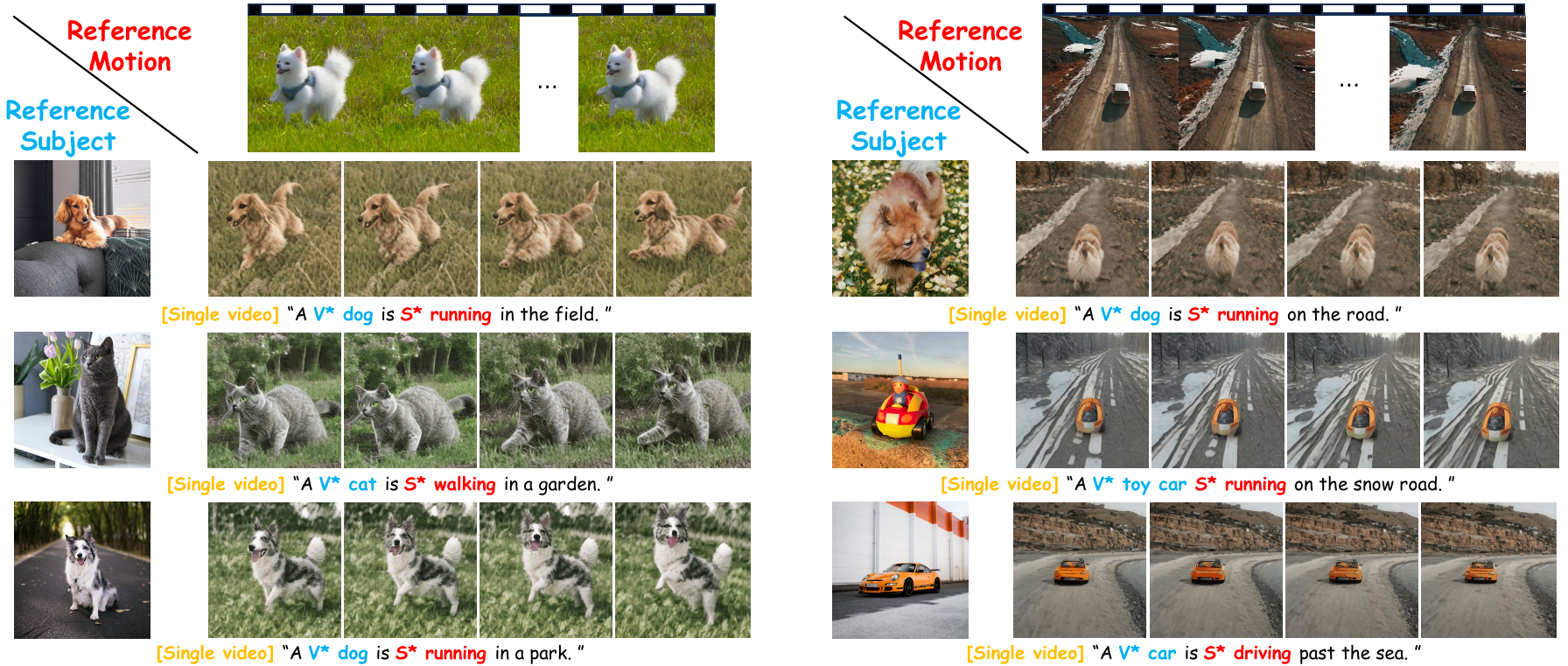}
     \vspace{-5pt}
    \caption{\textbf{More qualitative results of our joint customization for subject and motion.} In this case, we use a single video to guide our model in learning the specific object motion. SMRABooth generates customized videos that accurately preserve subject appearance and motion patterns while remaining faithful to text prompts.}
    \label{figconcept2}
   \vspace{-0.5cm}
\end{figure*}
\begin{figure*}[t]
    \centering
    \includegraphics[width=\textwidth]{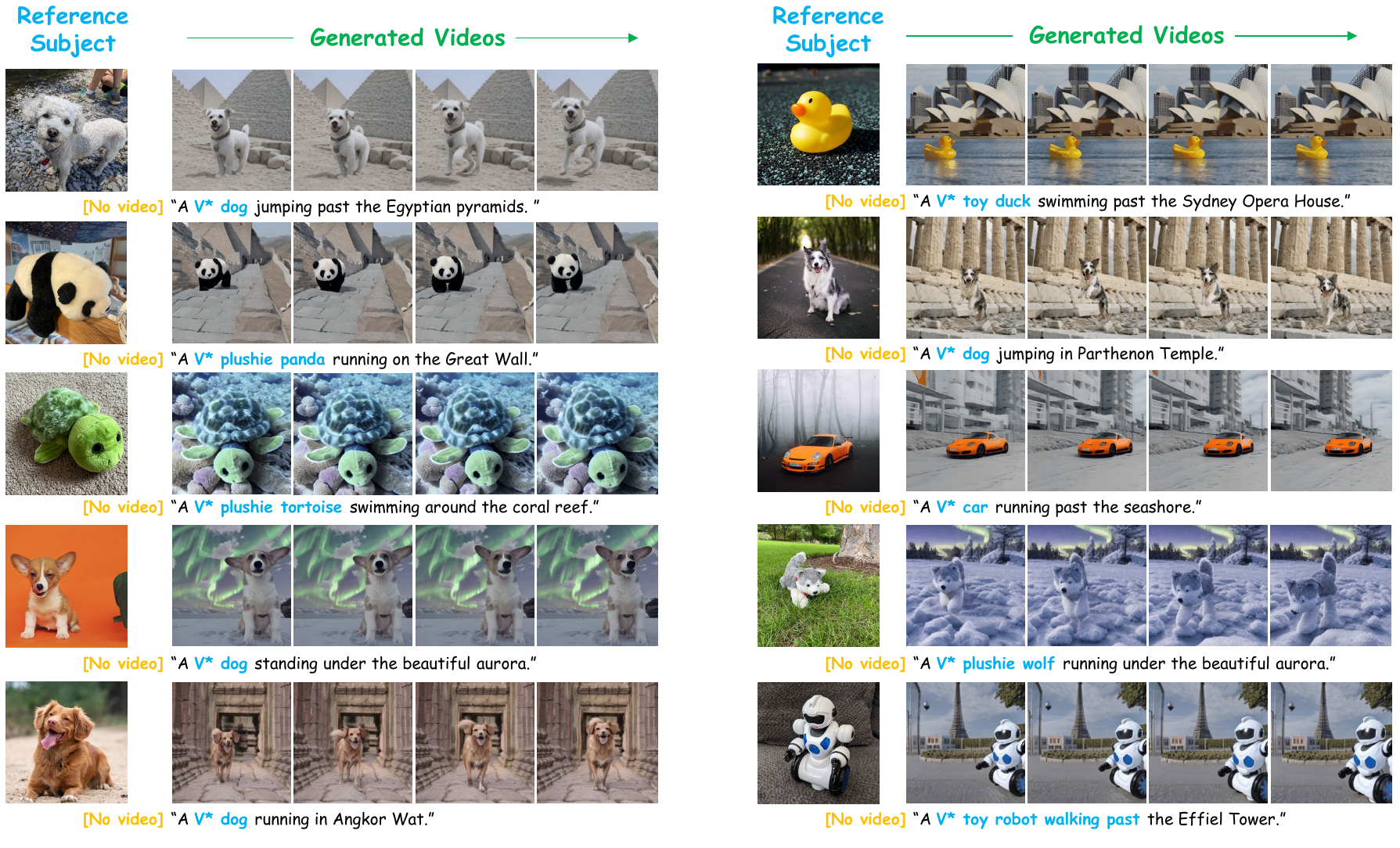}
    \caption{\textbf{More qualitative results of our customization for subject.} In this case, We have customized and generated a variety of different subjects, including various world historical sites and natural landscapes. Our cases fully demonstrate the accurate extraction of theme features and the strong generalization capabilities of our model.}
    \vspace{-0.2cm}
    \label{figconcept3}
\end{figure*}
\begin{figure*}[t]
    \centering
    \includegraphics[width=\textwidth]{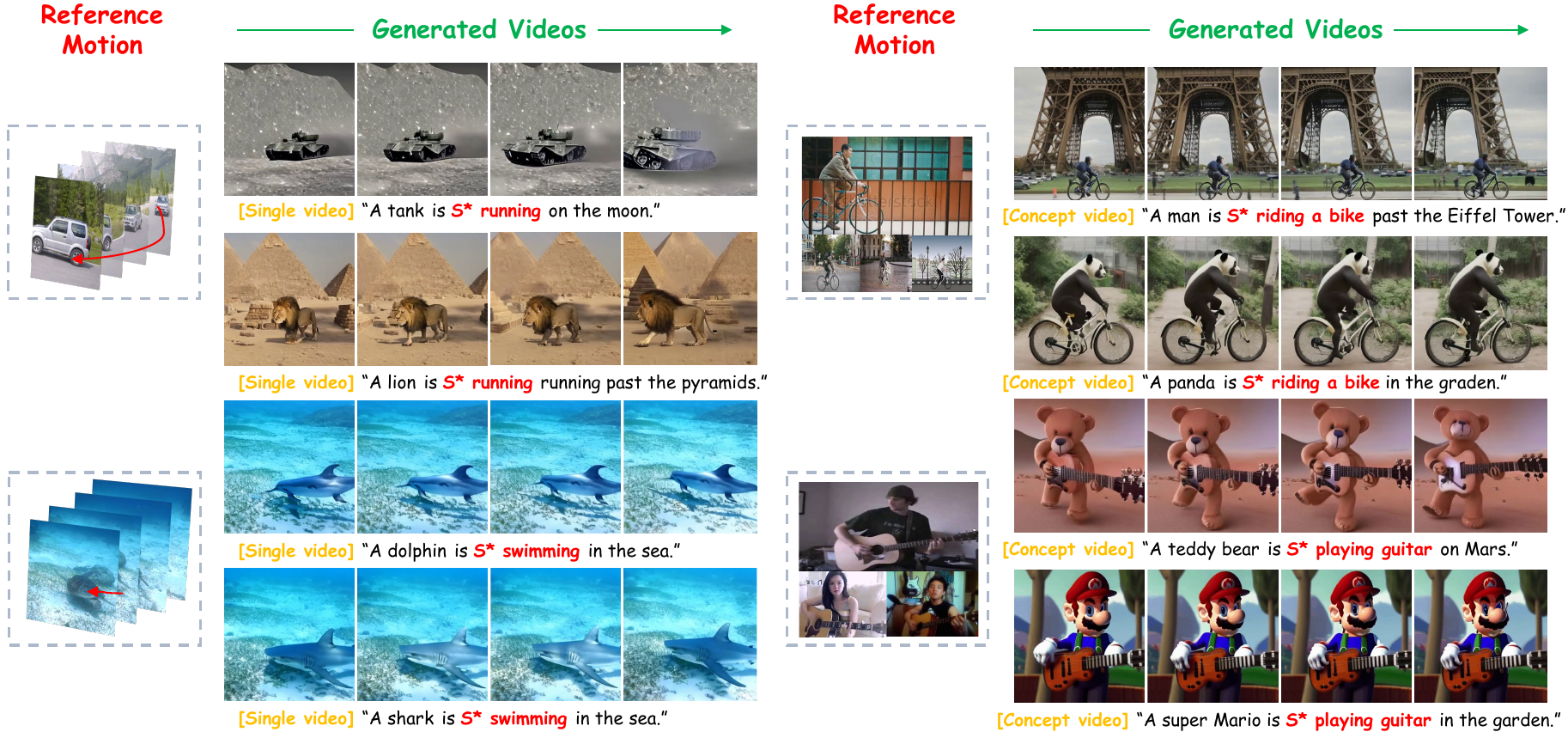}
    \caption{\textbf{More qualitative results of our customization for motion.} In this case, We have customized and generated a variety of different motion. Our cases fully demonstrate the accurate extraction of theme features and the strong generalization capabilities of our model.}
    \label{figconcept4}
    \vspace{-0.5cm}
\end{figure*}
\end{document}